\documentclass[conference]{IEEEtran}
\IEEEoverridecommandlockouts
\usepackage{cite}
\usepackage{amsmath,amssymb,amsfonts}
\usepackage{iftex}
\ifPDFTeX
  \usepackage[T1]{fontenc}
  \usepackage[utf8]{inputenc}
  \usepackage{mathptmx}
\else
  \usepackage{fontspec}
  \usepackage{unicode-math}
\fi
\usepackage{graphicx}
\usepackage{textcomp}
\usepackage{xcolor}
\usepackage{booktabs,array,multirow,longtable}
\usepackage{calc}
\usepackage{caption}
\usepackage{xurl}   
\usepackage[hidelinks]{hyperref}
\usepackage{etoolbox}

\AtBeginEnvironment{tabular}{\footnotesize}

\providecommand{\tightlist}{\setlength{\itemsep}{2pt}\setlength{\parskip}{0pt}}

\begin{document}

\title{A Large Open Multi-Energy Corpus of Soil Compaction Tests, with Machine-Learning Baselines}
\author{\IEEEauthorblockN{Sompote Youwai\IEEEauthorrefmark{1}, Chana Phutthananon, and Warat Kongkitkul}
\IEEEauthorblockA{AI Research Group, Department of Civil Engineering\\
King Mongkut's University of Technology Thonburi\\
126 Pracha Uthit Road, Bang Mod, Thung Khru, Bangkok 10140, Thailand\\
\IEEEauthorrefmark{1}Corresponding author: sompote.you@kmutt.ac.th}}
\maketitle

\begin{abstract}
Every engineered fill is specified by a maximum dry density and an optimum
moisture content. Each determination needs a full Proctor test. Published
correlations rest on one to four hundred specimens, usually from one laboratory
at one compactive energy, and are seldom released. This paper releases a corpus
without those limits. It holds 2,854 laboratory compaction tests from six public
sources, across 162 provenance groups and four Proctor energy levels, with fines
from 1.5 to 100 \%. Every record is audited to the Proctor method its source
names, and no energy is inferred. Screening on the zero-air-voids condition
removed 11.8 \% of harmonised records, and 5.7 \% of those with a measured
specific gravity. A material share of published compaction data is physically
impossible. The optimum degree of saturation over the corpus is 0.815 at a
coefficient of variation of 11 \%. That is a baseline, not a constant. Both
parameters are then estimated from one classification suite and the compaction
standard. A tabular foundation model reaches R\textsuperscript{2} 0.824 for density and 0.784 for
water content under random folds. It reaches 0.727 and 0.696 with folds drawn
around provenance, and 0.520 and 0.614 with a whole source held out. Compactive
energy is negligible marginally yet decisive conditionally. Density on the 66
modified-Proctor records is predicted at R\textsuperscript{2} 0.740 with it and $-$0.651 without.
Symbolic regression yields closed forms coupled through a phase relation. No
predicted pair can then exceed the zero-air-voids line. The predictions are for
screening, not acceptance.
\end{abstract}


\section{Introduction}\label{introduction}

\subsection{Compaction parameters and why they are expensive}\label{compaction-parameters-and-why-they-are-expensive}

Compaction control governs the strength, stiffness, compressibility and
permeability of every engineered fill. Two parameters specify it. The maximum
dry density (MDD) is the highest dry density reached under a stated compactive
effort. The optimum moisture content (OMC) is the water content at which that
maximum occurs. Both come from the test of Proctor {[}1{]}, standardised as ASTM
D698 {[}2{]} and D1557 {[}3{]} and as AASHTO T99 {[}4{]} and T180 {[}5{]}. The test compacts four to six
specimens at different water contents. Each is oven-dried for 24 h, so one
determination takes about a working day. Specifications require a determination
for every change of material. The number of tests therefore follows the
variability of the deposits, not the volume of fill placed. This burden
motivates the present study.

The optimum matters in its own right, not only as a companion to the density. It
governs soil fabric. Lambe {[}6{]} showed that a clay compacted dry of optimum
develops a flocculated particle arrangement, and the same soil compacted wet of
optimum develops a dispersed one. Seed and Chan {[}7{]} showed that the resulting
differences in strength, volume change and permeability persist at equal dry
density. The compaction water content also sets the soil-water potential of the
fill, and with it the susceptibility to swelling or to collapse on wetting
{[}8{]}. Daniel and Benson {[}9{]} proposed that acceptance be
defined as a zone in the dry density--water content plane rather than as
independent bounds on each quantity. That definition makes the position of the
optimum a design quantity. Continuous- and intelligent-compaction systems report
a mechanistic response across an entire lift {[}10{]}, and the
laboratory optimum remains the reference against which those measurements are
interpreted.

The optimum is bounded. The bound is the zero-air-voids (ZAV) line, along which
the remaining voids are entirely water-filled. No compactive effort can produce
a state above that line. The condition is therefore a criterion of physical
admissibility, and every record used here is screened against it (Section 2).
Raising the compactive effort raises MDD and reduces OMC. The peaks obtained at
different energy levels trace a locus approximately parallel to the ZAV line.
Analytical descriptions of that family of curves {[}11--13{]} start from one observation. For a given soil, the degree of
saturation at the optimum is nearly invariant.

Tatsuoka and Gomes Correia {[}14{]} developed that observation into a framework
for compaction control. They define the optimum degree of saturation
\((S_r)_{opt}\) as the saturation at which the maximum dry density is reached for
a given energy level. From a large body of laboratory and field data they show
that \((S_r)_{opt}\) and the normalised compaction curve change little with
limited variation in energy level and soil type. They argue that placement
should be specified in terms of degree of saturation rather than water content.
Tatsuoka et al.~{[}15{]} carry the same normalisation into field control, and the
nine soils tabulated there have \((S_r)_{opt}\) between 64.1 \% and 92.3 \%. Both
frameworks need \((\rho_d)_{max}\) and \((S_r)_{opt}\) for the material before any
field measurement can be interpreted. They therefore sit downstream of the
laboratory compaction test rather than in place of it. The present study
addresses the upstream problem of estimating those quantities for a material
that has not yet been compacted. The evidence behind those frameworks combines
laboratory and field data. The corpus assembled here is laboratory-only. The two
are not interchangeable, and Section 6 states the resulting restriction.

\subsection{Estimating the parameters from index properties}\label{estimating-the-parameters-from-index-properties}

Where the test is expensive and the parameters are needed for every material,
the established approach is to estimate them from properties already measured.
Correlations of that kind fall into two groups.

The first regresses on a single consistency limit. Sridharan and Nagaraj {[}16{]}
argued that the plastic limit governs the position of the peak. Gurtug and
Sridharan {[}17{]}, Nagaraj et al.~{[}18{]}, Sharma and Sridharan {[}19{]} and Vinod
and Sridharan {[}20{]} developed the resulting forms. Pandian et al.~{[}21{]} had
earlier re-examined the compaction characteristics of fine-grained soils on a
similar basis. Blotz et al.~{[}22{]} remain the standard reference for estimating
both parameters across compactive efforts from the liquid limit alone.

The second group applies regression or machine learning to a wider set of index
properties. It begins with the artificial neural networks of Günaydın {[}23{]} and
extends to recent tree-ensemble, deep-learning and hybrid formulations {[}24,25{]}. Verma and Sivapullaiah {[}26{]} review the
field. Reported coefficients of determination are frequently above 0.90.

Five features of the underlying datasets limit what those figures show. Table 1
lists them beside what this study does about each. The fifth is the feature this
paper is organised around. Almost every study in the second group scores its
model on a random hold-out or a shuffled k-fold partition of its own dataset.
Where that dataset comes from one laboratory, the resulting figure means what it
states. Where a dataset is assembled from several sources and then pooled and
shuffled, the meaning changes. A random fold can place records sharing a soil, a
pavement layer, a laboratory protocol or a publication on both sides of the
boundary. The score then measures interpolation within the assembled corpus.

{\def\LTcaptype{none} 
\begin{table*}[!tp]\centering
\caption*{Table 1 --- Five features of the datasets behind published compaction correlations, and what this study does about each.}
\begin{tabular}{@{}
>{\raggedright\arraybackslash}p{(\linewidth - 4\tabcolsep) * \real{0.4925}}
  >{\raggedright\arraybackslash}p{(\linewidth - 4\tabcolsep) * \real{0.2687}}
  >{\raggedright\arraybackslash}p{(\linewidth - 4\tabcolsep) * \real{0.2388}}@{}}
\toprule\noalign{}
\begin{minipage}[b]{\linewidth}\raggedright
Feature of the published datasets
\end{minipage} & \begin{minipage}[b]{\linewidth}\raggedright
What it prevents
\end{minipage} & \begin{minipage}[b]{\linewidth}\raggedright
Treatment here
\end{minipage} \\
\midrule\noalign{}
Size of 100--400 specimens, usually one laboratory & Coverage of the soils a project meets & 2,854 tests, six sources (Table 3) \\
One compactive energy on almost all records & Separating effort from the soils it was applied to & Four Proctor levels, energy never inferred (Table 6) \\
Data seldom released & Verification and transfer & Full corpus released with per-record provenance \\
Physical admissibility rarely examined & Knowing how much published data is impossible & Every record screened on the ZAV line (Table 5) \\
Scoring on a random hold-out of the study's own data & Knowing how the model behaves on a new source & Three fold designs, two of them transfer-facing (Table 9) \\
\bottomrule\noalign{}
\end{tabular}
\end{table*}

}

\subsection{What public compaction data exists}\label{what-public-compaction-data-exists}

Two kinds of dataset underlie the correlations above. Neither is a database in
the sense used here.

The first kind is the set a study assembles for its own paper. These are the one
to four hundred specimens already described, and they are rarely deposited. A
correlation keyed on a consistency limit is also undefined for a soil that has
none, so that branch of the literature is restricted to plastic material by
construction. Its reference points are accordingly clays and fine-grained soils
{[}17,21,22{]}. The largest
compilation in this literature is that of Rehman et al.~{[}25{]}. They assemble
1,001 observations and model them across compactive energy levels with
gradient-boosted trees, reporting the liquid limit and the energy as the two
leading predictors. That work is the closest antecedent of the present study and
the point from which it departs. Its scope is cohesive soils, as its title
states. A granular fill has no consistency limits and therefore no place in it.
The compilation is not among the deposits retained in Section 2.

The second kind is the deposited dataset. Section 2 reports a census of these
rather than a sample, and Table 2 gives the result. The retained deposits are
small, between 7 and 395 records. Each comes from one laboratory or one
compilation at one compactive energy. Beside them sits the Long-Term Pavement
Performance release, which is large at 1,976 records but is a single testing
programme following one protocol at standard effort. No public source holds the
individual points of a compaction curve, and none holds a field density.

What is assembled here is therefore not a larger version of any of them. It is
the six sources pooled under one set of harmonisation conventions. The
provenance of every record is kept, so a fold can be drawn around it. Four
compactive energies are present rather than one. 134 non-plastic soils are
carried alongside clays of liquid limit above 100 \%. The median fines content is
48.6 \% against a range of 1.5 to 100 \%. The corpus therefore sits across the
coarse--fine boundary rather than on the cohesive side of it.

\subsection{Scope and objectives}\label{scope-and-objectives}

The consequence for practice is direct. An engineer meeting a new borrow
material cannot tell whether a published correlation was fitted on soils like
it. The engineer can neither refit it nor know how it will behave on material
from a laboratory the correlation never saw. The gap is therefore not a better
regression on the datasets that already exist. It is the absence of the data
those regressions need.

Pooling the sources does three things that none of them does alone. It spans the
soils a project actually meets rather than one deposit. It carries more than one
compactive energy, so effort becomes a variable rather than a constant. And it
keeps the provenance of every record, so a model can be scored on a laboratory
it has never seen. That last point is the question an engineer is asking, and it
governs the evaluation throughout.

Two estimators are then given, because they are used differently. A fitted
ensemble is the more accurate. It has to be run as software, its coefficients
cannot be inspected, and nothing in it prevents a physically impossible
prediction. A closed-form expression is less accurate. It can be written into a
specification, checked by hand, and built so that the pair it returns cannot lie
above the zero-air-voids line. Both are reported, and the cost of the closed
form is measured rather than assumed.

The estimation problem is stated in full in Section 4.1 and summarised in
Table 8. One routine classification suite on a disturbed sample gives the
plastic limit, the plasticity index, the fines content, the sand content and the
specific gravity. The compaction standard is also known. The task is to estimate
MDD and OMC from those inputs. No quantity derived from the compaction curve
itself is used as a predictor.

This study has three objectives.

\begin{enumerate}
\def\labelenumi{\arabic{enumi}.}
\item
  To assemble a corpus of laboratory compaction tests from public sources. It
  holds 2,854 tests from six sources, spans four Proctor energy levels and
  crosses the coarse--fine boundary. It is an order of magnitude larger than the
  datasets on which the published correlations rest. Every record carries the
  compactive energy its source states, and records failing the zero-air-voids
  condition are reported rather than removed silently.
\item
  To evaluate a model that predicts MDD and OMC from a single classification
  suite. Three classes of learner are carried throughout, so the results reflect
  the database rather than one modelling choice. The evaluation uses fold
  designs that respect the provenance of each record.
\item
  To propose closed-form expressions for both parameters. The optimum moisture
  content is recovered from the estimated density through the phase relation at
  a fixed optimum degree of saturation. That constraint keeps every estimated
  pair below the zero-air-voids line.
\end{enumerate}

Three further results follow. Drawing folds around the 162 provenance groups
costs about 0.10 in R\textsuperscript{2} at an identical training-set size. Compactive energy
appears negligible marginally, yet it is decisive once plasticity and gradation
are held constant. The optimum degree of saturation of Tatsuoka and Gomes
Correia {[}14{]}, measured over 2,854 tests, has a mean of 0.815 and a coefficient
of variation of 11 \%.

Section 2 assembles and screens the corpus. Section 3 describes the surviving
records. Section 4 defines the inputs, the models and the three fold designs.
Section 5 reports the results. Section 6 states what the corpus cannot support,
and Section 7 concludes.

\section{Assembling and screening the corpus}\label{assembling-and-screening-the-corpus}

\subsection{Search and retention}\label{search-and-retention}

Two channels were searched in August 2026, following PRISMA 2020 {[}27{]} and PRISMA-S {[}28{]}. Table 2 gives the result of each.

Deposited datasets were sought through the DataCite application programming
interface across Zenodo, figshare, Dryad, Mendeley Data and the Open Science
Framework. DataCite registers most research datasets, so this search
approximates a census rather than a sample. Open-access articles were located
through OpenAlex. Both channels reach only open-access material and query in
English, so the corpus samples the openly available literature rather than
compaction testing at large.

A deposited dataset was retained only if it met five conditions: an open licence
permitting redistribution, both MDD and OMC on every record, at least one index
property on every record, provenance identifiable at record level, and the two
targets being the peak of a laboratory compaction test. The last condition sets
the scope of everything that follows. Field densities, nuclear-gauge readings
and optima back-calculated from field data are all outside it.

{\def\LTcaptype{none} 
\begin{table*}[!tp]\centering
\caption*{Table 2 --- Search and retention. Both channels were run in August 2026. The article stream was excluded on completeness, not on quality.}
\begin{tabular}{@{}
>{\raggedright\arraybackslash}p{(\linewidth - 6\tabcolsep) * \real{0.2812}}
  >{\raggedleft\arraybackslash}p{(\linewidth - 6\tabcolsep) * \real{0.1875}}
  >{\raggedleft\arraybackslash}p{(\linewidth - 6\tabcolsep) * \real{0.1042}}
  >{\raggedright\arraybackslash}p{(\linewidth - 6\tabcolsep) * \real{0.4271}}@{}}
\toprule\noalign{}
\begin{minipage}[b]{\linewidth}\raggedright
Channel
\end{minipage} & \begin{minipage}[b]{\linewidth}\raggedleft
Screened
\end{minipage} & \begin{minipage}[b]{\linewidth}\raggedleft
Retained
\end{minipage} & \begin{minipage}[b]{\linewidth}\raggedright
Reason for exclusion
\end{minipage} \\
\midrule\noalign{}
DataCite deposits & 618 candidates & 5 & licence, missing target, or no record-level provenance \\
--- of which, integrity & 2 deposits, 502 records & 0 & near-duplicate records perturbed in the fourth to sixth decimal \\
--- of which, curve points & all candidates & 0 & no repository holds individual compaction-curve points \\
OpenAlex articles and theses & 11,119 documents, 4,370 parsed, 2,746 records & 0 & fines on 3.7 \%, sand on 4.8 \%, energy on none \\
LTPP Standard Data Release 39 & 1 programme & 1 & --- \\
\textbf{Total retained} & & \textbf{6 sources} & \\
\bottomrule\noalign{}
\end{tabular}
\end{table*}

}

Two exclusions deserve comment because they are decisions rather than filters.
Two figshare deposits of 502 records met all five conditions but were excluded
on integrity. Each follows every apparently genuine specimen with about nine
near-copies perturbed in the fourth to sixth decimal place. The article and
thesis stream was excluded in full on completeness. Extraction recovers both
targets on all of it, but fines on 3.7 \% of records, sand on 4.8 \% and
compactive energy on none. Admitting it is the most costly choice reported
anywhere in this paper. It takes the gradient-boosted model from R\textsuperscript{2} 0.819 to
0.536 for density, and from 0.783 to 0.467 for water content.

\subsection{The six retained sources}\label{the-six-retained-sources}

Table 3 describes the six sources. Every retained record is the peak of a
laboratory compaction curve, and the origin of each was audited source by
source.

For LTPP the evidence is documentary. The programme's own documentation
describes the source table as holding standard Proctor results {[}29{]}, which makes every LTPP record AASHTO T99 at 592.5 kJ/m\textsuperscript{3}.
The density field in that table carries no unit, and only one of the two
candidate readings returns physically sensible saturations, which settles it.
For the five deposits the evidence is the deposit itself. No source reports a
field density, so the dataset is laboratory-only. Every statement in this paper
about MDD, OMC and \((S_r)_{opt}\) is a statement about laboratory compaction.

{\def\LTcaptype{none} 
\begin{table*}[!tp]\centering
\caption*{Table 3 --- The six retained sources. Deposits are named by their identifier: LTPP SDR 39 is Federal Highway Administration {[}30{]}, figshare 28681187 is Günaydın {[}23{]}, Zenodo 20737270 is the Türkiye deposit and Zenodo 14251190 is Soranzo {[}31{]}. \(n\) is records retained after screening. The fines and OMC columns are medians over those records. \(G_s\) meas. counts the records reporting a measured specific gravity rather than the substituted median.}
\begin{tabular}{@{}
>{\raggedright\arraybackslash}p{(\linewidth - 14\tabcolsep) * \real{0.1979}}
  >{\raggedleft\arraybackslash}p{(\linewidth - 14\tabcolsep) * \real{0.0625}}
  >{\raggedleft\arraybackslash}p{(\linewidth - 14\tabcolsep) * \real{0.0521}}
  >{\raggedright\arraybackslash}p{(\linewidth - 14\tabcolsep) * \real{0.1875}}
  >{\raggedright\arraybackslash}p{(\linewidth - 14\tabcolsep) * \real{0.3125}}
  >{\raggedleft\arraybackslash}p{(\linewidth - 14\tabcolsep) * \real{0.0625}}
  >{\raggedleft\arraybackslash}p{(\linewidth - 14\tabcolsep) * \real{0.0625}}
  >{\raggedleft\arraybackslash}p{(\linewidth - 14\tabcolsep) * \real{0.0625}}@{}}
\toprule\noalign{}
\begin{minipage}[b]{\linewidth}\raggedright
Source
\end{minipage} & \begin{minipage}[b]{\linewidth}\raggedleft
\(n\)
\end{minipage} & \begin{minipage}[b]{\linewidth}\raggedleft
Groups
\end{minipage} & \begin{minipage}[b]{\linewidth}\raggedright
A group is
\end{minipage} & \begin{minipage}[b]{\linewidth}\raggedright
Laboratory method
\end{minipage} & \begin{minipage}[b]{\linewidth}\raggedleft
Fines (\%)
\end{minipage} & \begin{minipage}[b]{\linewidth}\raggedleft
OMC (\%)
\end{minipage} & \begin{minipage}[b]{\linewidth}\raggedleft
\(G_s\) meas.
\end{minipage} \\
\midrule\noalign{}
LTPP SDR 39 & 1,976 & 54 & a state & AASHTO T99, standard & 45.2 & 12.0 & 935 \\
figshare 28681187 & 395 & 5 & the original reference & standard Proctor & 38.0 & 13.8 & 0 \\
Zenodo 20737270 & 269 & 1 & undivided & standard Proctor & 66.7 & 19.0 & 0 \\
figshare 32955851 & 169 & 83 & the original reference & ASTM D698 and D1557, four levels & 60.0 & 15.0 & 169 \\
Zenodo 14251190 & 38 & 12 & the site & Proctor, standard not named & 84.9 & 18.7 & 38 \\
Zenodo 19242689 & 7 & 7 & the soil & standard Proctor & 95.0 & 22.0 & 6 \\
\textbf{Total} & \textbf{2,854} & \textbf{162} & & & \textbf{48.6} & \textbf{13.0} & \textbf{1,148} \\
\bottomrule\noalign{}
\end{tabular}
\end{table*}

}

Two things follow from Table 3. LTPP carries 69 \% of the records, and its
material is coarser and drier than four of the five deposits. A pooled accuracy
figure is therefore weighted towards LTPP soils. And the multi-energy deposit,
figshare 32955851, is a single source. Every non-standard test in the corpus
comes from it, which constrains what the source-held-out design can test
(Sections 5.4 and 6).

\subsection{Harmonisation}\label{harmonisation}

Three columns need not mean the same thing in every source. Each was audited
record by record, and Table 4 states the convention adopted and its measured
consequence.

{\def\LTcaptype{none} 
\begin{table*}[!tp]\centering
\caption*{Table 4 --- The three harmonisation audits. The effect column reports a measurement, not an assumption.}
\begin{tabular}{@{}
>{\raggedright\arraybackslash}p{(\linewidth - 6\tabcolsep) * \real{0.1284}}
  >{\raggedright\arraybackslash}p{(\linewidth - 6\tabcolsep) * \real{0.2936}}
  >{\raggedright\arraybackslash}p{(\linewidth - 6\tabcolsep) * \real{0.2844}}
  >{\raggedright\arraybackslash}p{(\linewidth - 6\tabcolsep) * \real{0.2936}}@{}}
\toprule\noalign{}
\begin{minipage}[b]{\linewidth}\raggedright
Column
\end{minipage} & \begin{minipage}[b]{\linewidth}\raggedright
The problem
\end{minipage} & \begin{minipage}[b]{\linewidth}\raggedright
Convention adopted
\end{minipage} & \begin{minipage}[b]{\linewidth}\raggedright
Measured effect
\end{minipage} \\
\midrule\noalign{}
Particle size & ASTM D2487 {[}32{]} divides gravel from sand at 4.75 mm; AASHTO and European systems at 2.00 mm, and most sources do not state theirs & LTPP fractions recomputed at 4.75 mm; other records carry their source's boundary & Restating LTPP at 2.00 mm moves MDD R\textsuperscript{2} from 0.789 to 0.794, inside the fold-to-fold dispersion \\
Compactive energy & Sources name a standard, not a value & No value read from a source; energy follows from the blow-count specification of the named standard & Four levels: 355, 593, 1,347 and 2,693 kJ/m\textsuperscript{3} \\
Specific gravity & 59.8 \% of retained records report none & Reported value used where given; corpus median of measured values, 2.68, substituted and flagged otherwise & Nearly inert as an input; decisive in the screening rule (Table 5) \\
\bottomrule\noalign{}
\end{tabular}
\end{table*}

}

The energy convention is worth stating twice, because it is what makes the
corpus multi-energy. No compactive energy is inferred from the data. A record
whose source does not identify its Proctor method is excluded rather than
assigned a value. The four levels are reduced standard, standard, reduced
modified and modified Proctor.

\subsection{Physical admissibility}\label{physical-admissibility}

Admissibility was checked on every harmonised record, using quantities the
record already contains. With \(\rho_w\) taken as 1 Mg/m\textsuperscript{3},

\begin{align}
e &= \frac{G_s}{\mathrm{MDD}} - 1 \label{eq:void-ratio}\\[2pt]
S_{opt} &= \frac{\mathrm{OMC}\cdot G_s}{e} \label{eq:s-opt}
\end{align}

\(S_{opt}\) is the optimum degree of saturation \((S_r)_{opt}\) of Tatsuoka and
Gomes Correia {[}14{]}, computed here per record rather than per soil. It gives a
direct test. A value above unity puts the record above the zero-air-voids line,
which is impossible as published. A very low value points to a quantity that has
been scaled wrongly. Records were kept where \(S_{opt}\) lies between 0.20 and
1.0, MDD between 0.8 and 2.6 Mg/m\textsuperscript{3} and OMC between 1 and 70 \%. The window on
\(S_{opt}\) is deliberately wide, because the quantity is derived and inherits any
error in an assumed specific gravity.

Table 5 gives the result. Screening removed 11.8 \% of the 8,242 harmonised
records, and 94 \% of those removals were above the zero-air-voids line. Taken at
face value, about one published compaction record in nine describes a soil that
cannot exist.

{\def\LTcaptype{none} 
\begin{table*}[!tp]\centering
\caption*{Table 5 --- The zero-air-voids screen. The upper block splits the harmonised records by how specific gravity was obtained. The lower block re-runs the screen on the same records with the substituted value changed.}
\begin{tabular}{@{}
>{\raggedright\arraybackslash}p{(\linewidth - 8\tabcolsep) * \real{0.5574}}
  >{\raggedleft\arraybackslash}p{(\linewidth - 8\tabcolsep) * \real{0.1148}}
  >{\raggedleft\arraybackslash}p{(\linewidth - 8\tabcolsep) * \real{0.1148}}
  >{\raggedleft\arraybackslash}p{(\linewidth - 8\tabcolsep) * \real{0.0984}}
  >{\raggedleft\arraybackslash}p{(\linewidth - 8\tabcolsep) * \real{0.1148}}@{}}
\toprule\noalign{}
\begin{minipage}[b]{\linewidth}\raggedright
Subset or setting
\end{minipage} & \begin{minipage}[b]{\linewidth}\raggedleft
Records
\end{minipage} & \begin{minipage}[b]{\linewidth}\raggedleft
Removed
\end{minipage} & \begin{minipage}[b]{\linewidth}\raggedleft
Rate (\%)
\end{minipage} & \begin{minipage}[b]{\linewidth}\raggedleft
Above ZAV
\end{minipage} \\
\midrule\noalign{}
All harmonised records, \(G_s\) as published or substituted & 8,242 & 969 & 11.8 & 914 \\
--- records with a measured \(G_s\) & 2,737 & 156 & 5.7 & 156 \\
--- records with a substituted \(G_s\) & 5,505 & 813 & 14.8 & 758 \\
Substituted value set to 2.60 & 8,242 & 1,630 & 19.8 & 1,575 \\
Substituted value set to 2.65 & 8,242 & 1,161 & 14.1 & 1,106 \\
Substituted value set to 2.68 (adopted) & 8,242 & 969 & 11.8 & 914 \\
Substituted value set to 2.70 & 8,242 & 863 & 10.5 & 807 \\
Substituted value set to 2.75 & 8,242 & 678 & 8.2 & 622 \\
\bottomrule\noalign{}
\end{tabular}
\end{table*}

}

The two blocks of Table 5 answer different questions. The upper block shows that
the rejection rate is more than twice as high where specific gravity had to be
substituted. The lower block is the more reliable comparison, because the
records do not change between runs. Moving the substituted value from 2.60 to
2.75 moves the rate from 19.8 \% to 8.2 \%. The finding that a material share of
published compaction data is physically impossible is robust. The figure of one
in nine is not.

The full protocol behind this section is released with the code, including the
query strings, the per-candidate screening decisions, the source adapters and
the three harmonisation audits.

\section{The dataset}\label{the-dataset}

\subsection{Composition and the two deliberate gaps}\label{composition-and-the-two-deliberate-gaps}

The records surviving the screening of Section 2 form the dataset modelled here.
It holds 2,854 laboratory compaction tests across 162 provenance groups. A
record is included when it carries every field the model requires, which is a
test of model-readiness rather than of completeness. Table 6 gives the
composition by compactive effort and the state of every column.

{\def\LTcaptype{none} 
\begin{table*}[!tp]\centering
\caption*{Table 6 --- Composition of the modelling set, \(n\) = 2,854. The upper block is by compactive effort. The lower block reports the columns that are not complete, and what an absent entry means in each.}
\begin{tabular}{@{}
llrrr@{}}
\toprule\noalign{}
Compactive effort & Standard & Energy (kJ/m\textsuperscript{3}) & Records & Share (\%) \\
\midrule\noalign{}
Reduced standard & ASTM D698 A, 15 blows & 355 & 28 & 1.0 \\
Standard & ASTM D698 A / AASHTO T99 & 593 & 2,753 & 96.5 \\
Reduced modified & ASTM D1557 A, half effort & 1,347 & 7 & 0.2 \\
Modified & ASTM D1557 A / AASHTO T180 & 2,693 & 66 & 2.3 \\
\bottomrule\noalign{}
\end{tabular}

\vspace{5pt}

\begin{tabular}{@{}
>{\raggedright\arraybackslash}p{(\linewidth - 8\tabcolsep) * \real{0.2135}}
  >{\raggedleft\arraybackslash}p{(\linewidth - 8\tabcolsep) * \real{0.0787}}
  >{\raggedleft\arraybackslash}p{(\linewidth - 8\tabcolsep) * \real{0.0674}}
  >{\raggedright\arraybackslash}p{(\linewidth - 8\tabcolsep) * \real{0.2921}}
  >{\raggedright\arraybackslash}p{(\linewidth - 8\tabcolsep) * \real{0.3483}}@{}}
\toprule\noalign{}
\begin{minipage}[b]{\linewidth}\raggedright
Incomplete column
\end{minipage} & \begin{minipage}[b]{\linewidth}\raggedleft
Absent on
\end{minipage} & \begin{minipage}[b]{\linewidth}\raggedleft
Share (\%)
\end{minipage} & \begin{minipage}[b]{\linewidth}\raggedright
What absence means
\end{minipage} & \begin{minipage}[b]{\linewidth}\raggedright
How it reaches the model
\end{minipage} \\
\midrule\noalign{}
Liquid and plastic limit & 134 & 4.7 & not applicable: the soil is non-plastic & passed to the learner as a gap \\
Sand content & 169 & 5.9 & not reported: the multi-energy source publishes fines but no grading curve & passed to the learner as a gap \\
Specific gravity & 1,706 & 59.8 & not reported: corpus median 2.68 substituted and flagged & substituted value plus a flag \\
\bottomrule\noalign{}
\end{tabular}
\end{table*}

}

The two gaps are kept rather than filled, and the reason differs for each.
Requiring sand would delete every record from figshare 32955851 and so fix
energy at a single level, which removes the variable of principal interest. The
consistency limits are missing where the tests do not apply, so an absent value
is the correct entry rather than a number. Most of those 134 records also report
a plastic limit above the liquid limit, which cannot be a measurement. XGBoost
and TabPFN take both gaps directly. The perceptron cannot, so it receives the
training-fold median together with a was-missing indicator, and the indicator
carries the information.

\subsection{Provenance and what a random fold leaks}\label{provenance-and-what-a-random-fold-leaks}

Every record carries the identifier of the publication, site or deposit it came
from. That gives 162 groups, and Table 7 describes them. The groups are very
uneven: 94 hold a single record and the largest holds 269.

{\def\LTcaptype{none} 
\begin{table*}[!tp]\centering
\caption*{Table 7 --- The grouping variable, and what a random five-fold partition places on both sides of the boundary.}
\begin{tabular}{@{}
>{\raggedright\arraybackslash}p{(\linewidth - 2\tabcolsep) * \real{0.8889}}
  >{\raggedleft\arraybackslash}p{(\linewidth - 2\tabcolsep) * \real{0.1111}}@{}}
\toprule\noalign{}
\begin{minipage}[b]{\linewidth}\raggedright
Property
\end{minipage} & \begin{minipage}[b]{\linewidth}\raggedleft
Value
\end{minipage} \\
\midrule\noalign{}
Provenance groups & 162 \\
Groups holding a single record & 94 \\
Records in the largest group & 269 \\
Records whose input vector is repeated at least once & 1,405 \\
Repetitions that lie inside a single group & all \\
Under a random fold: records whose own group is in the training fold & 96.6 \% \\
Under a random fold: records whose exact input vector is in the training fold & 40.1 \% \\
\bottomrule\noalign{}
\end{tabular}
\end{table*}

}

Two properties of this variable govern the evaluation. First, records sharing a
group share much more than a label. They share a soil deposit, a laboratory, a
sampling campaign and a reporting convention, and they are often repeat tests on
one pavement layer. Under a random partition the model is therefore handed a
near-copy, and in 40 \% of cases an exact copy, of the record it is about to be
scored on. Second, the grouping is coarse where the corpus is heaviest. An LTPP
group is a state rather than a laboratory, and all 54 belong to one testing
programme following one protocol. Holding out a state tests transfer to a new
material, not to a new laboratory or a new reporting convention.

\subsection{The dataset in the compaction plane}\label{the-dataset-in-the-compaction-plane}

Figure 1 shows the dataset. Panel (a) places every record on the MDD--OMC plane,
marked by the compactive effort its source states, with the zero-air-voids line
and two constant-saturation contours at \(G_s\) = 2.68. Panel (b) gives the six
sources side by side on five quantities.

\begin{figure*}[!tp]\centering
\includegraphics[width=\textwidth]{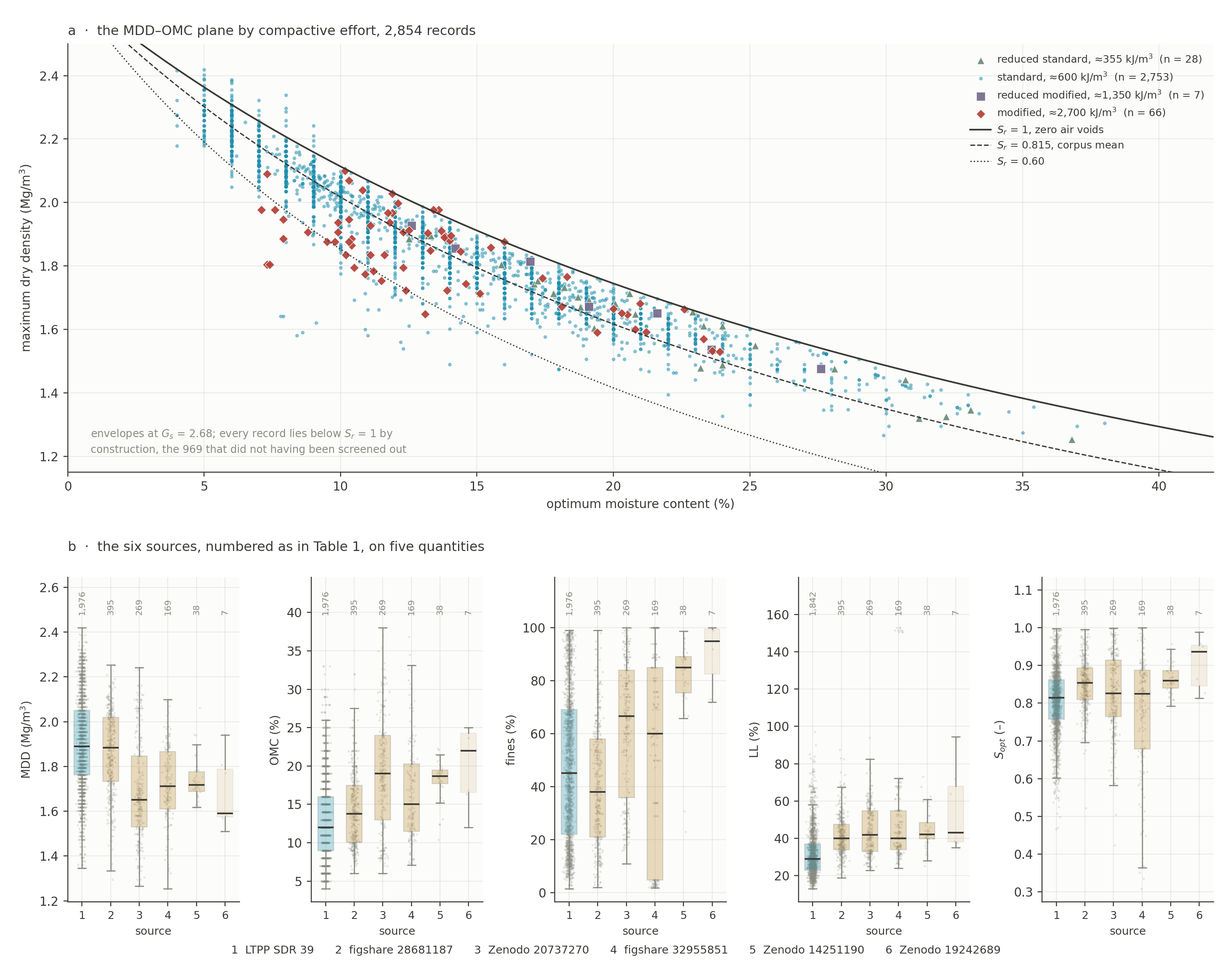}
\caption{The dataset: (a) the MDD–OMC plane by compactive effort, with envelopes at $G_s$ = 2.68; (b) the six sources of Table 3 on five quantities.}
\end{figure*}

Three features of panel (a) matter later. The corpus occupies a band beneath the
ZAV line rather than a cloud. A soil reaches its peak at an approximately fixed
degree of saturation, so MDD and OMC are two coordinates of one point on a
saturation curve. The band is closed above by the screening rule of Section 2
and thins below \(S_r\) = 0.6. The modified-effort records do not sit above the
standard ones. Between 5 and 10 \% water content their median density is 0.192
Mg/m\textsuperscript{3} below the standard records at the same water content, because the soils
compacted harder are also the more plastic. The vertical striping reflects LTPP
reporting the optimum to the nearest whole per cent, which accounts for many of
the repeated input vectors of Table 7.

Three consequences follow for the rest of the paper. The band structure means
the two targets are not independent quantities to be estimated separately, which
is why Section 5.5 recovers the optimum from the estimated density through the
phase relation. The overlap of the effort classes means compactive energy cannot
be read from the plane alone, so Section 5.4 scores each effort level separately
rather than pooling them. And the weight of LTPP within the corpus means Section
4.3 must draw folds around provenance.

\section{Methodology}\label{methodology}

\subsection{The estimation problem}\label{the-estimation-problem}

Section 3 established two properties the methodology must respect. The corpus
carries a group structure, and two of its columns are deliberately incomplete.
The estimation problem is stated once against that background, and Table 8 sets
it out.

The inputs are five index properties from a single routine classification suite
on a disturbed sample, together with the compaction standard to which the
material is to be compacted. From those the task is to estimate the two
coordinates of the laboratory compaction-curve peak. No input is required beyond
what a classification and a specific-gravity determination already produce. No
quantity taken from the compaction curve itself enters the predictors.

{\def\LTcaptype{none} 
\begin{table*}[!tp]\centering
\caption*{Table 8 --- Inputs, targets, and the quantities the corpus holds but the mapping never sees. Index properties are reported as percentages. OMC is modelled as a decimal fraction and reported as a percentage.}
\begin{tabular}{@{}
>{\raggedright\arraybackslash}p{(\linewidth - 8\tabcolsep) * \real{0.1013}}
  >{\raggedright\arraybackslash}p{(\linewidth - 8\tabcolsep) * \real{0.3038}}
  >{\raggedright\arraybackslash}p{(\linewidth - 8\tabcolsep) * \real{0.1139}}
  >{\raggedright\arraybackslash}p{(\linewidth - 8\tabcolsep) * \real{0.3924}}
  >{\raggedleft\arraybackslash}p{(\linewidth - 8\tabcolsep) * \real{0.0886}}@{}}
\toprule\noalign{}
\begin{minipage}[b]{\linewidth}\raggedright
Symbol
\end{minipage} & \begin{minipage}[b]{\linewidth}\raggedright
Quantity
\end{minipage} & \begin{minipage}[b]{\linewidth}\raggedright
Unit
\end{minipage} & \begin{minipage}[b]{\linewidth}\raggedright
Obtained from
\end{minipage} & \begin{minipage}[b]{\linewidth}\raggedleft
Absent on
\end{minipage} \\
\midrule\noalign{}
\(w_P\) & plastic limit & \% & Atterberg limits, ASTM D4318 {[}33{]} & 134 \\
\(I_P\) & plasticity index & \% & \(w_L - w_P\), arithmetic & 134 \\
\(F\) & fines content & \% & sieve and hydrometer & 0 \\
\(s\) & sand content & \% & sieve and hydrometer & 169 \\
\(G_s\) & specific gravity & -- & pycnometer, ASTM D854 {[}34{]} & 1,706 substituted \\
\(\ln \bar{E}\) & log compactive energy & -- & the compaction standard, declared not measured & 0 \\
\bottomrule\noalign{}
\end{tabular}

\vspace{5pt}

\begin{tabular}{@{}
lll@{}}
\toprule\noalign{}
Target & Quantity & Unit \\
\midrule\noalign{}
MDD & maximum dry density & Mg/m\textsuperscript{3} \\
OMC & optimum moisture content & \% \\
\bottomrule\noalign{}
\end{tabular}

\vspace{5pt}

\begin{tabular}{@{}
>{\raggedright\arraybackslash}p{(\linewidth - 2\tabcolsep) * \real{0.4342}}
  >{\raggedright\arraybackslash}p{(\linewidth - 2\tabcolsep) * \real{0.5658}}@{}}
\toprule\noalign{}
\begin{minipage}[b]{\linewidth}\raggedright
Held out of the mapping
\end{minipage} & \begin{minipage}[b]{\linewidth}\raggedright
Why
\end{minipage} \\
\midrule\noalign{}
\((S_r)_{opt}\) & derived from both targets; using it would leak them \\
liquid limit \(w_L\) & the third of three limits is the difference of the other two (Table 13) \\
source identifier & drives grouping and stratification; a new material has no place in the corpus \\
\bottomrule\noalign{}
\end{tabular}
\end{table*}

}

Figure 2 sets the problem out as five stages: what a laboratory measures, the
arithmetic on those measurements, the mapping itself, the pair of targets, and
the intended use.

\begin{figure*}[!tp]\centering
\includegraphics[width=0.5\textwidth]{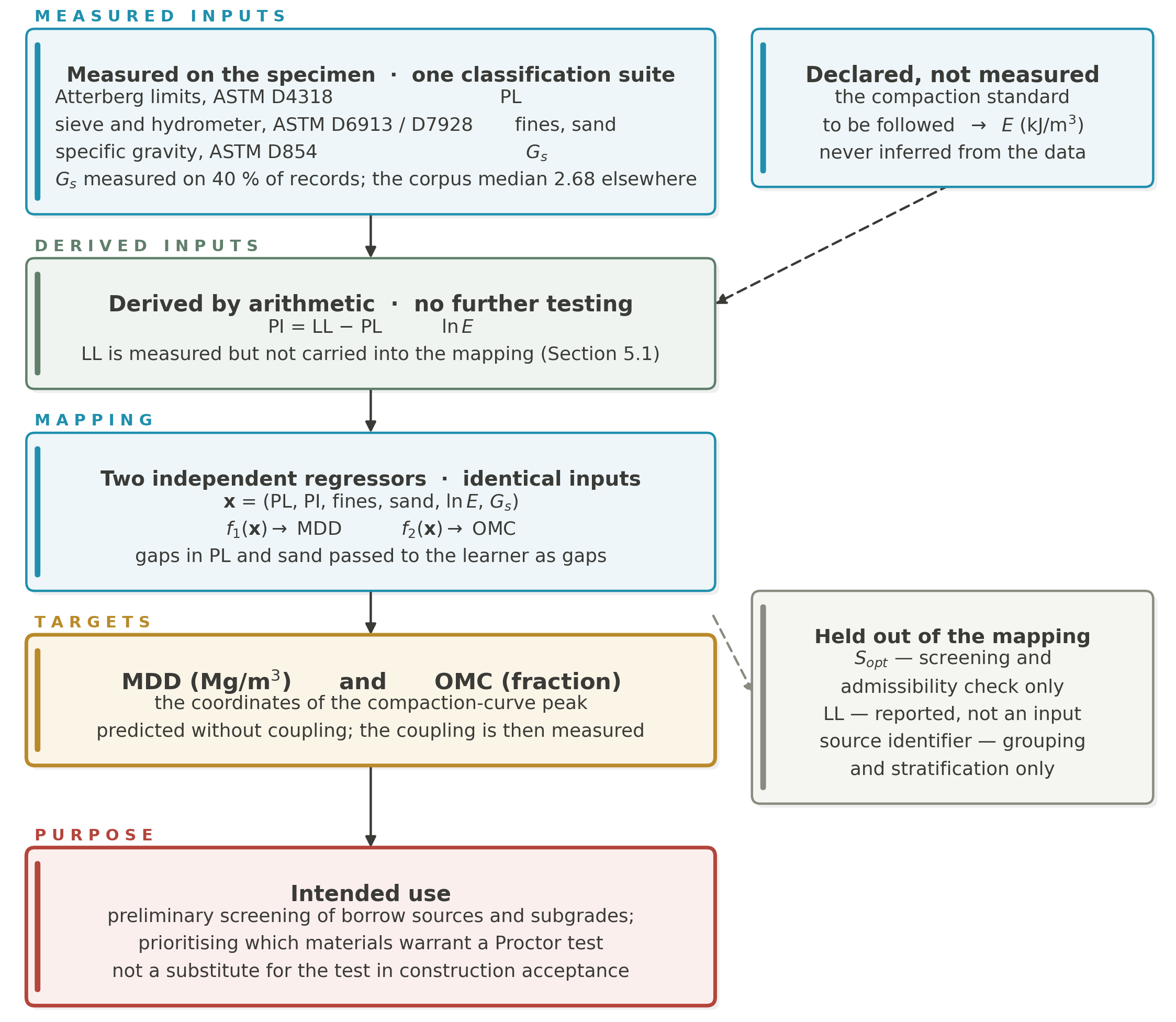}
\caption{The estimation problem: six inputs, two targets, and the quantities held out of the mapping.}
\end{figure*}

Formally, writing \(\mathbf{x} = (w_P, I_P, F, s, \ln\bar{E}, G_s) \in
\mathbb{R}^{6}\), two regressors are fitted independently, \(f_1(\mathbf{x})
\rightarrow \mathrm{MDD}\) and \(f_2(\mathbf{x}) \rightarrow \mathrm{OMC}\), on
identical inputs and identical partitions. The two targets are physically
coupled, being the coordinates of a single curve peak, but no coupling is
imposed on the estimators. Whether they nevertheless respect it is measured in
Section 5.1, and a structurally coupled alternative is given in Section 5.5.

\textbf{Intended use.} Three uses are intended: ranking borrow sources before any is
tested, deciding which material warrants a Proctor determination first, and
setting the water content of its first specimen. They are not a basis for
accepting placed fill. A mean absolute error of 0.066 Mg/m\textsuperscript{3} is about 3.5 \% of
the median maximum dry density. That is the same order as the margin a
specification leaves when it requires 95 \% of MDD in the field, so an acceptance
decision would spend that entire margin on the prediction. The estimate is
inexpensive not because the six inputs are free, but because they are made in
any case and made once per material, whereas a Proctor test is made once per
material and compactive standard.

\subsection{Models}\label{models}

Three learners carry the principal results. Eight further architectures drawn
from the compaction literature are refitted on identical folds in Appendix A,
which tests at breadth the proposition that the accuracy reached is a property
of the data rather than of the model class. Section 5.5 additionally searches
for closed-form expressions by symbolic regression using PySR {[}35,36{]}.

TabPFN {[}37{]} predicts by in-context learning. The training
records are supplied as context and the test records evaluated in a single
forward pass, with no gradient step on the present data. The network is
pre-trained once, on synthetic datasets drawn from structural causal models, to
approximate the posterior predictive distribution under that prior {[}38{]}. It therefore introduces no hyperparameter and no search conducted on
these records, so no comparison below can be confounded by unequal search
effort. The whole training fold is re-supplied for every prediction, which is
why the cost is 522 s against 2 s for the ensemble.

The gradient-boosted model is an \texttt{XGBRegressor} with values fixed in advance. A
nested random search over nine hyperparameters, selected on a 3-fold inner
cross-validation within each training fold, returned R\textsuperscript{2} 0.816 and 0.779 against
0.819 and 0.783 for the fixed values. Tuning therefore yields no improvement.
The perceptron and the two transformer baselines are described with the released
code.

\subsection{The three fold designs}\label{the-three-fold-designs}

The boundary around which a fold is drawn determines what the resulting score
means. Where records carry a group structure, a fold that ignores it reports a
score for interpolation within the groups rather than prediction for a new one.
The remedy is to block the folds on the grouping variable {[}39{]}.
Three designs are used here, on the same 2,854 records, the same six inputs and
the same learners. Only the boundary changes between them. Table 9 defines them.

{\def\LTcaptype{none} 
\begin{table*}[!tp]\centering
\caption*{Table 9 --- The three fold designs. The random and grouped designs are matched in training-set size and differ only in composition.}
\begin{tabular}{@{}
>{\raggedright\arraybackslash}p{(\linewidth - 10\tabcolsep) * \real{0.1591}}
  >{\raggedright\arraybackslash}p{(\linewidth - 10\tabcolsep) * \real{0.2159}}
  >{\raggedleft\arraybackslash}p{(\linewidth - 10\tabcolsep) * \real{0.0455}}
  >{\raggedleft\arraybackslash}p{(\linewidth - 10\tabcolsep) * \real{0.1364}}
  >{\raggedleft\arraybackslash}p{(\linewidth - 10\tabcolsep) * \real{0.0682}}
  >{\raggedright\arraybackslash}p{(\linewidth - 10\tabcolsep) * \real{0.3750}}@{}}
\toprule\noalign{}
\begin{minipage}[b]{\linewidth}\raggedright
Design
\end{minipage} & \begin{minipage}[b]{\linewidth}\raggedright
Boundary
\end{minipage} & \begin{minipage}[b]{\linewidth}\raggedleft
Folds
\end{minipage} & \begin{minipage}[b]{\linewidth}\raggedleft
Training records
\end{minipage} & \begin{minipage}[b]{\linewidth}\raggedleft
Test record shares its group with training
\end{minipage} & \begin{minipage}[b]{\linewidth}\raggedright
Question it answers
\end{minipage} \\
\midrule\noalign{}
Random & none, shuffled at a fixed seed & 5 & 2,283 & 96.6 \% & how well the corpus interpolates within itself \\
Grouped & the 162 provenance groups & 5 & 2,283--2,284 & 0 \% & transfer to a publication, state dataset or site not seen in training \\
Source held out & the six sources of Table 3 & 6 & 878--2,847 & 0 \% & transfer to a whole data source not seen in training \\
\bottomrule\noalign{}
\end{tabular}
\end{table*}

}

The random design is the convention of the comparative literature. It is
retained rather than replaced, because it is the design under which the figures
below may be set beside published ones, and because the difference between it
and the grouped design is itself a result. It is not the basis of any statement
about a new laboratory, publication or data source.

The source-held-out design is the harshest of the three. Holding out LTPP leaves
878 training records, and holding out any other source leaves a training set at
least 69 \% LTPP. A fourth design, leave-one-group-out over all 162 groups, was
also run for the gradient-boosted model and is released with the code. Its error
distribution across groups is skewed: the median group is predicted at 0.066
Mg/m\textsuperscript{3} and the ninetieth percentile at 0.110, but the tail runs to 0.268.

Performance is reported as the coefficient of determination, the mean absolute
error and the root-mean-square error, against the floor set by a mean predictor
computed on the training fold and scored on the identical held-out records. Each
is computed once over the pooled out-of-fold predictions rather than as the mean
of per-fold scores, because the admissibility check requires one prediction
attached to every record. Fold-to-fold dispersion is reported separately as the
$\pm$ in Table 10. Under the source-held-out design the coefficient of determination
for a source is taken against that source's own mean, which is a harder
reference than the corpus mean where a source is narrow, so the mean absolute
error is set beside it. Because MDD and OMC are predicted separately, a predicted
pair may be physically impossible, and the proportion above the zero-air-voids
line is reported as a consistency check. Bold in a results table marks the best
value in the comparison that table makes, and a note beneath each table states
which comparison that is.

Three qualifications apply. First, the symbolic search of Section 5.5 runs on a
single 80/20 random partition. The evolutionary search sees only the 2,176
training records, but two later steps are not held to the same standard: the
entry taken off the Pareto front was chosen with held-out accuracy in view, and
the constant \((S_r)_{opt}\) is a mean over all 2,720 records. Both are model
selection performed with sight of the 544, so the accuracies quoted for that
partition are selection scores rather than test scores. The expressions are
refitted and rescored under all three designs in Section 5.5. Second, no
untouched test set exists, and Section 6 states what that costs. Third, the
1,405 repeated input vectors of Table 7 cannot leak under the grouped or
source-held-out designs, since every repetition lies inside one group. Under the
random design they can, and the consequence was measured. Collapsing every
repeated vector to a single record carrying the median target leaves 2,126
records and changes the random result by $-$0.005 in R\textsuperscript{2} for MDD and +0.002 for
OMC. That effect is small while the grouped design costs an order of magnitude
more, so duplication is not the mechanism. Shared provenance is.

\section{Results}\label{results}

\subsection{Predictive accuracy}\label{predictive-accuracy}

Table 10 reports the three learners on the six inputs under all three fold
designs, so that the leakage the random design admits can be read against the
figure it produces. Within a design, every learner is trained and scored on one
fold assignment.

TabPFN attains R\textsuperscript{2} 0.824 for maximum dry density at a mean absolute error of
0.066 Mg/m\textsuperscript{3}, and 0.784 for optimum moisture content at 1.87 \% water content.
These are reductions in error of 63 \% and 60 \% against the mean predictor. Both
leading learners lie within the range at which the prediction is of practical
use. An error of 1.87 \% in optimum moisture content is comparable with the
between-laboratory reproducibility of the Proctor test, and 0.066 Mg/m\textsuperscript{3} is below
4 \% of the median density.

{\def\LTcaptype{none} 
\begin{table*}[!tp]\centering
\caption*{Table 10 --- Out-of-fold performance, six inputs, 2,854 records. The last column is a mean predictor computed on the training fold and scored on the identical held-out records. Water-content errors are percentage points. The $\pm$ is the standard deviation across the five random folds. The last block counts predicted pairs implying a degree of saturation above unity.}
\begin{tabular}{@{}
>{\raggedright\arraybackslash}p{(\linewidth - 8\tabcolsep) * \real{0.6596}}
  >{\raggedleft\arraybackslash}p{(\linewidth - 8\tabcolsep) * \real{0.0851}}
  >{\raggedleft\arraybackslash}p{(\linewidth - 8\tabcolsep) * \real{0.0851}}
  >{\raggedleft\arraybackslash}p{(\linewidth - 8\tabcolsep) * \real{0.0851}}
  >{\raggedleft\arraybackslash}p{(\linewidth - 8\tabcolsep) * \real{0.0851}}@{}}
\toprule\noalign{}
\begin{minipage}[b]{\linewidth}\raggedright
Design and metric
\end{minipage} & \begin{minipage}[b]{\linewidth}\raggedleft
\textbf{TabPFN}
\end{minipage} & \begin{minipage}[b]{\linewidth}\raggedleft
XGBoost
\end{minipage} & \begin{minipage}[b]{\linewidth}\raggedleft
Perceptron
\end{minipage} & \begin{minipage}[b]{\linewidth}\raggedleft
Mean
\end{minipage} \\
\midrule\noalign{}
\textbf{Random folds} & & & & \\
MDD R\textsuperscript{2} & \textbf{0.824 $\pm$ 0.012} & 0.819 $\pm$ 0.014 & 0.791 $\pm$ 0.017 & $-$0.000 \\
MDD MAE (Mg/m\textsuperscript{3}) & \textbf{0.066 $\pm$ 0.002} & 0.068 $\pm$ 0.002 & 0.074 $\pm$ 0.002 & 0.178 \\
MDD RMSE (Mg/m\textsuperscript{3}) & \textbf{0.091} & 0.093 & 0.099 & 0.217 \\
OMC R\textsuperscript{2} & \textbf{0.784 $\pm$ 0.006} & 0.783 $\pm$ 0.011 & 0.756 $\pm$ 0.003 & $-$0.001 \\
OMC MAE (\%) & \textbf{1.87 $\pm$ 0.08} & 1.88 $\pm$ 0.07 & 2.03 $\pm$ 0.10 & 4.61 \\
OMC RMSE (\%) & \textbf{2.68} & 2.69 & 2.85 & 5.77 \\
\textbf{Grouped folds} & & & & \\
MDD R\textsuperscript{2} & \textbf{0.727} & 0.722 & 0.717 & $-$0.023 \\
MDD MAE (Mg/m\textsuperscript{3}) & \textbf{0.086} & 0.088 & 0.087 & 0.180 \\
OMC R\textsuperscript{2} & \textbf{0.696} & 0.681 & 0.690 & $-$0.024 \\
OMC MAE (\%) & \textbf{2.28} & 2.36 & 2.32 & 4.66 \\
\textbf{Source held out} & & & & \\
MDD R\textsuperscript{2} & 0.520 & 0.392 & \textbf{0.577} & $-$0.235 \\
MDD MAE (Mg/m\textsuperscript{3}) & 0.116 & 0.127 & \textbf{0.110} & 0.196 \\
OMC R\textsuperscript{2} & \textbf{0.614} & 0.225 & 0.555 & $-$0.258 \\
OMC MAE (\%) & \textbf{2.76} & 3.53 & 2.82 & 5.32 \\
\textbf{Pairs above the ZAV line}, \(n\) & & & & \\
random folds & \textbf{1} & 7 & 12 & 28 \\
grouped folds & \textbf{0} & 38 & 23 & -- \\
source held out & 41 & \textbf{8} & 27 & -- \\
\bottomrule\noalign{}
\end{tabular}
\end{table*}

}

\emph{Bold marks the best value in each row. The mean-predictor column is a floor,
not a competitor, and is excluded from the comparison.}

Three readings follow from Table 10, and they are separate.

\emph{The random figures are interpolation figures.} They are the only figures in
this paper that can be set beside published ones. They measure how well the
corpus interpolates within itself, not how well the model transfers to a source
it has not seen. Drawing the fold boundary around the provenance group, at an
identical training-set size, costs about 0.10 in R\textsuperscript{2} on both targets and on every
learner. The mean absolute error in water content rises from 1.87 to 2.28 \%.
Nothing about the model, the inputs or the amount of training data changes
between those two blocks. What changes is whether the model had already seen the
publication on which it was scored.

\emph{The learners agree under two designs and separate under the third.} Under
random and grouped folds the three span 0.03 in density. With a whole source held
out they span 0.19, and the ordering reverses. Appendix A shows that the cause is
the missing-data encoding rather than the architecture.

\emph{Accuracy is not the only criterion.} Because the two targets are predicted
independently, a predicted pair may imply a degree of saturation above unity.
The learners separate more sharply on that criterion than on accuracy. Predicting
each target by its training-fold mean already places 28 pairs above the line, so
the perceptron does no better than a constant under random folds. The rate cannot
be driven to zero by improved fitting, because nothing in the formulation forbids
a violation. Building the phase relation into the formulation does forbid it,
which is the approach of Section 5.5.

Sections 5.4 and 5.5 are reported with TabPFN and the gradient-boosted model. On
this evidence neither is more accurate than the other, so the choice rests on two
other grounds. TabPFN puts 1 of its 2,854 random-fold pairs above the ZAV line
against 7 for the ensemble, a sevenfold difference on a quantity that is not
measurement noise. And TabPFN has no hyperparameters, so no result below can be
attributed to a search conducted on these records.

\subsection{Transfer to an unseen source}\label{transfer-to-an-unseen-source}

Table 11 opens the source-held-out block of Table 10 source by source. Each
source in turn is predicted by a model trained on the other five. The
coefficient of determination is taken against that source's own mean, so a
narrow source is scored against a hard reference. The standard deviation of the
target within each source is given for that reason.

{\def\LTcaptype{none} 
\begin{table*}[!tp]\centering
\caption*{Table 11 --- Leave-one-source-out, source by source. R\textsuperscript{2} is taken against the held-out source's own mean. sd is the standard deviation of the target within that source, given because it sets how hard that reference is.}
\begin{tabular}{@{}
>{\raggedright\arraybackslash}p{(\linewidth - 12\tabcolsep) * \real{0.3519}}
  >{\raggedleft\arraybackslash}p{(\linewidth - 12\tabcolsep) * \real{0.0926}}
  >{\raggedleft\arraybackslash}p{(\linewidth - 12\tabcolsep) * \real{0.1111}}
  >{\raggedleft\arraybackslash}p{(\linewidth - 12\tabcolsep) * \real{0.1111}}
  >{\raggedleft\arraybackslash}p{(\linewidth - 12\tabcolsep) * \real{0.1111}}
  >{\raggedleft\arraybackslash}p{(\linewidth - 12\tabcolsep) * \real{0.1111}}
  >{\raggedleft\arraybackslash}p{(\linewidth - 12\tabcolsep) * \real{0.1111}}@{}}
\toprule\noalign{}
\begin{minipage}[b]{\linewidth}\raggedright
Held-out source
\end{minipage} & \begin{minipage}[b]{\linewidth}\raggedleft
\(n\)
\end{minipage} & \begin{minipage}[b]{\linewidth}\raggedleft
sd
\end{minipage} & \begin{minipage}[b]{\linewidth}\raggedleft
TabPFN R\textsuperscript{2}
\end{minipage} & \begin{minipage}[b]{\linewidth}\raggedleft
TabPFN MAE
\end{minipage} & \begin{minipage}[b]{\linewidth}\raggedleft
XGBoost R\textsuperscript{2}
\end{minipage} & \begin{minipage}[b]{\linewidth}\raggedleft
XGBoost MAE
\end{minipage} \\
\midrule\noalign{}
\textbf{Maximum dry density} (sd and MAE in Mg/m\textsuperscript{3}) & & & & & & \\
LTPP SDR 39 & 1,976 & 0.210 & \textbf{0.436} & \textbf{0.121} & 0.286 & 0.130 \\
figshare 28681187 & 395 & 0.181 & \textbf{0.500} & \textbf{0.103} & 0.459 & 0.108 \\
Zenodo 20737270 & 269 & 0.223 & \textbf{0.664} & \textbf{0.105} & 0.648 & 0.107 \\
figshare 32955851 & 169 & 0.173 & \textbf{0.216} & \textbf{0.122} & $-$0.490 & 0.179 \\
Zenodo 14251190 & 38 & 0.096 & 0.196 & \textbf{0.060} & \textbf{0.278} & 0.061 \\
Zenodo 19242689 & 7 & 0.159 & $-$0.443 & 0.159 & \textbf{$-$0.164} & \textbf{0.148} \\
\textbf{Optimum moisture content} (sd and MAE in percentage points) & & & & & & \\
LTPP SDR 39 & 1,976 & 5.14 & \textbf{0.577} & \textbf{2.62} & $-$0.050 & 3.52 \\
figshare 28681187 & 395 & 4.81 & \textbf{0.507} & \textbf{2.67} & 0.462 & 2.78 \\
Zenodo 20737270 & 269 & 7.34 & 0.501 & \textbf{3.97} & \textbf{0.506} & 3.97 \\
figshare 32955851 & 169 & 6.47 & \textbf{0.678} & \textbf{2.87} & 0.130 & 5.01 \\
Zenodo 14251190 & 38 & 2.65 & \textbf{0.154} & \textbf{1.70} & $-$0.014 & 1.76 \\
Zenodo 19242689 & 7 & 5.06 & $-$0.644 & 5.39 & \textbf{$-$0.250} & \textbf{4.59} \\
\bottomrule\noalign{}
\end{tabular}
\end{table*}

}

\emph{Bold marks the better of the two learners in each row. On Zenodo 20737270 the
two water-content errors differ only in the third decimal.}

Three things are visible in Table 11 that the pooled figure hides. The hardest
fold for the gradient-boosted model is figshare 32955851 at R\textsuperscript{2} $-$0.490 for
density. That is the multi-energy deposit, and removing it removes every
non-standard record from training, leaving the energy variable present but
untrained at three of its four levels. The smallest source, at 7 records, is
predicted worse than its own mean by both learners, so no per-source conclusion
should be drawn from it. And the two learners diverge most on the OMC column of
the two largest folds, which is where the missing-data encoding of Appendix A
does its damage.

Leave-one-group-out over all 162 groups gives a complementary view of the same
question at finer resolution. The median group is predicted at a mean absolute
error of 0.066 Mg/m\textsuperscript{3} in density and 1.73 \% in water content. The ninetieth
percentile is 0.110 Mg/m\textsuperscript{3} and 3.49 \%, and the worst group reaches 0.268 Mg/m\textsuperscript{3}
and 11.0 \%. The error distribution across groups is therefore strongly skewed,
and a pooled mean absolute error understates the risk on an unlucky material.

\subsection{What the estimate rests on}\label{what-the-estimate-rests-on}

Three questions are asked of the input set. Which input groups carry the
prediction, which of the three consistency limits should be used, and what the
continuous prediction adds to a soil-classification label.

\textbf{Input groups.} Table 12 removes each input group in turn from the
gradient-boosted model and rescores under the random and grouped designs.

{\def\LTcaptype{none} 
\begin{table*}[!tp]\centering
\caption*{Table 12 --- Input ablation, gradient-boosted model. $\Delta$ is the loss in R\textsuperscript{2} against the full six-input model in the first row. Gradation is fines and sand together.}
\begin{tabular}{@{}
>{\raggedright\arraybackslash}p{(\linewidth - 8\tabcolsep) * \real{0.3684}}
  >{\raggedleft\arraybackslash}p{(\linewidth - 8\tabcolsep) * \real{0.1579}}
  >{\raggedleft\arraybackslash}p{(\linewidth - 8\tabcolsep) * \real{0.1579}}
  >{\raggedleft\arraybackslash}p{(\linewidth - 8\tabcolsep) * \real{0.1579}}
  >{\raggedleft\arraybackslash}p{(\linewidth - 8\tabcolsep) * \real{0.1579}}@{}}
\toprule\noalign{}
\begin{minipage}[b]{\linewidth}\raggedright
Inputs
\end{minipage} & \begin{minipage}[b]{\linewidth}\raggedleft
Random MDD R\textsuperscript{2} ($\Delta$)
\end{minipage} & \begin{minipage}[b]{\linewidth}\raggedleft
Random OMC R\textsuperscript{2} ($\Delta$)
\end{minipage} & \begin{minipage}[b]{\linewidth}\raggedleft
Grouped MDD R\textsuperscript{2} ($\Delta$)
\end{minipage} & \begin{minipage}[b]{\linewidth}\raggedleft
Grouped OMC R\textsuperscript{2} ($\Delta$)
\end{minipage} \\
\midrule\noalign{}
All six (reference) & 0.819 & 0.783 & 0.722 & 0.681 \\
Without the consistency limits & 0.650 (0.169) & 0.576 (0.207) & 0.534 (0.188) & 0.468 (0.213) \\
Without gradation entirely & 0.679 (0.140) & 0.686 (0.096) & 0.512 (0.210) & 0.540 (0.142) \\
Without sand only & 0.795 (0.024) & 0.771 (0.012) & 0.692 (0.030) & 0.670 (0.012) \\
Without compactive energy & 0.794 (0.025) & 0.761 (0.022) & 0.707 (0.015) & 0.664 (0.017) \\
Without specific gravity & 0.805 (0.013) & 0.775 (0.008) & 0.713 (0.009) & 0.673 (0.008) \\
\bottomrule\noalign{}
\end{tabular}
\end{table*}

}

Table 12 separates the inputs into two tiers. The consistency limits and
gradation each cost between 0.096 and 0.213 in R\textsuperscript{2}. Sand alone, compactive energy
and specific gravity never cost more than 0.030, so the cheapest entry in the
leading tier is still more than three times the most expensive in the second.
Which of the two leading groups leads depends on the target and the design, so
they are best read as jointly load-bearing rather than ranked. Gradation earns
its place through the fines content rather than through sand: removing both
grading terms costs seven times what removing sand alone does under grouped
folds. The two leading costs both grow when the fold boundary respects
provenance, so neither is inflated by shared provenance. The three small costs do
not change materially between designs, and the pooled energy figure is qualified
in Section 5.4.

\textbf{Consistency limits.} Of the three limits only two are algebraically
independent, since \(I_P = w_L - w_P\). Table 13 scores all four sets under the
three designs.

{\def\LTcaptype{none} 
\begin{table*}[!tp]\centering
\caption*{Table 13 --- The four consistency-limit sets, gradient-boosted model, each with fines, sand, energy and specific gravity. Each cell gives R\textsuperscript{2} for density then water content.}
\begin{tabular}{@{}
>{\raggedright\arraybackslash}p{(\linewidth - 6\tabcolsep) * \real{0.3390}}
  >{\raggedright\arraybackslash}p{(\linewidth - 6\tabcolsep) * \real{0.2203}}
  >{\raggedright\arraybackslash}p{(\linewidth - 6\tabcolsep) * \real{0.2203}}
  >{\raggedright\arraybackslash}p{(\linewidth - 6\tabcolsep) * \real{0.2203}}@{}}
\toprule\noalign{}
\begin{minipage}[b]{\linewidth}\raggedright
Limit set
\end{minipage} & \begin{minipage}[b]{\linewidth}\raggedright
Random
\end{minipage} & \begin{minipage}[b]{\linewidth}\raggedright
Grouped
\end{minipage} & \begin{minipage}[b]{\linewidth}\raggedright
Source held out
\end{minipage} \\
\midrule\noalign{}
all three, \(w_L\), \(w_P\), \(I_P\) & 0.818 / 0.781 & 0.713 / 0.679 & 0.351 / 0.162 \\
\(w_L\) and \(w_P\) & 0.818 / 0.782 & 0.719 / 0.671 & 0.252 / 0.062 \\
\(w_L\) and \(I_P\) & 0.810 / 0.767 & 0.701 / 0.673 & 0.120 / $-$0.043 \\
\textbf{\(w_P\) and \(I_P\) (used here)} & \textbf{0.819 / 0.783} & \textbf{0.722 / 0.681} & \textbf{0.392 / 0.225} \\
\bottomrule\noalign{}
\end{tabular}
\end{table*}

}

\emph{Bold marks the highest value in each column. It is the set used throughout, and
it is best under all three designs and on both targets.}

The pair \(w_P\) with \(I_P\) equals or exceeds all three limits together in every
design. The margin is negligible under random folds and 0.041 in density with a
source held out. The three-limit set is not rank-deficient, but the redundant
column costs transfer. The liquid limit is therefore released with the dataset
and not carried into the mapping. Section 6 records that this choice was made on
the learners and charged to the closed forms, which prefer \(w_L\).

\textbf{Value beyond a classification label.} Gradation and the consistency limits
already place a soil in a classification group, and a group carries a qualitative
expectation of how it will compact. Table 14 asks what the continuous prediction
adds to that label. Every class mean is computed inside the training fold and
scored on the held-out records, so the label is evaluated on the same terms as
the model.

{\def\LTcaptype{none} 
\begin{table*}[!tp]\centering
\caption*{Table 14 --- What the prediction adds to a classification label. Each cell gives R\textsuperscript{2} for density then water content. The USCS block covers all 2,854 records. The AASHTO block covers the 1,976 LTPP records, the only ones carrying an AASHTO symbol.}
\begin{tabular}{@{}
>{\raggedright\arraybackslash}p{(\linewidth - 8\tabcolsep) * \real{0.4091}}
  >{\raggedright\arraybackslash}p{(\linewidth - 8\tabcolsep) * \real{0.1477}}
  >{\raggedright\arraybackslash}p{(\linewidth - 8\tabcolsep) * \real{0.1477}}
  >{\raggedright\arraybackslash}p{(\linewidth - 8\tabcolsep) * \real{0.1477}}
  >{\raggedright\arraybackslash}p{(\linewidth - 8\tabcolsep) * \real{0.1477}}@{}}
\toprule\noalign{}
\begin{minipage}[b]{\linewidth}\raggedright
Predictor
\end{minipage} & \begin{minipage}[b]{\linewidth}\raggedright
USCS, random
\end{minipage} & \begin{minipage}[b]{\linewidth}\raggedright
USCS, grouped
\end{minipage} & \begin{minipage}[b]{\linewidth}\raggedright
AASHTO, random
\end{minipage} & \begin{minipage}[b]{\linewidth}\raggedright
AASHTO, grouped
\end{minipage} \\
\midrule\noalign{}
Corpus mean & $-$0.000 / $-$0.001 & $-$0.023 / $-$0.024 & $-$0.001 / $-$0.001 & $-$0.013 / $-$0.013 \\
Class mean & 0.563 / 0.561 & 0.543 / 0.524 & 0.626 / 0.616 & 0.611 / 0.604 \\
Six inputs, ordinary least squares & 0.679 / 0.696 & 0.630 / 0.675 & 0.735 / 0.716 & 0.732 / 0.712 \\
Six inputs, gradient-boosted trees & 0.819 / 0.783 & 0.722 / 0.681 & 0.796 / 0.753 & 0.713 / 0.669 \\
\bottomrule\noalign{}
\end{tabular}
\end{table*}

}

The USCS symbol alone explains 0.563 of the variance in density and 0.561 in
water content under random folds. The six inputs used linearly add 0.116 and
0.134 on top of that. Used by an ensemble they add a further 0.139 and 0.086.
Under grouped folds the margin narrows without closing. A class mean is a
low-capacity predictor that was never exploiting shared provenance, so it falls
only to 0.543 while the ensemble falls to 0.722. The gain over the label is then
0.179 rather than 0.256 in density, and 0.157 rather than 0.222 in water content.
The practical advantage survives the change of design at about two-thirds of its
apparent size.

Two causes separate the label from the prediction. A classification group is a
broad interval. CL spans liquid limits from 20 to 50 \% and fines from 50 to
100 \%, so soils that compact differently share a symbol. And the residual gap
between the linear model and the ensemble is the cost of assuming additivity.
Neither is removed by classifying more finely. The AASHTO block of Table 14 uses
a finer scheme on the LTPP records and both gaps persist.

\subsection{Effect of compactive energy}\label{effect-of-compactive-energy}

Table 12 ranked compactive energy near the bottom of the six inputs, at 0.015 to
0.025 in R\textsuperscript{2}. That figure substantially understates its influence, because
standard Proctor accounts for 96.5 \% of the dataset. The pooled value is an
average over a corpus in which the variable barely varies. Table 15 splits the
same predictions by the effort actually applied, and Figure 3 draws them.

{\def\LTcaptype{none} 
\begin{table*}[!tp]\centering
\caption*{Table 15 --- R\textsuperscript{2} and MAE for density within each effort stratum, with and without the energy input, gradient-boosted model. The strata are those of Table 6.}
\begin{tabular}{@{}
>{\raggedright\arraybackslash}p{(\linewidth - 10\tabcolsep) * \real{0.2500}}
  >{\raggedleft\arraybackslash}p{(\linewidth - 10\tabcolsep) * \real{0.0833}}
  >{\raggedright\arraybackslash}p{(\linewidth - 10\tabcolsep) * \real{0.1667}}
  >{\raggedright\arraybackslash}p{(\linewidth - 10\tabcolsep) * \real{0.1667}}
  >{\raggedright\arraybackslash}p{(\linewidth - 10\tabcolsep) * \real{0.1667}}
  >{\raggedright\arraybackslash}p{(\linewidth - 10\tabcolsep) * \real{0.1667}}@{}}
\toprule\noalign{}
\begin{minipage}[b]{\linewidth}\raggedright
Effort
\end{minipage} & \begin{minipage}[b]{\linewidth}\raggedleft
\(n\)
\end{minipage} & \begin{minipage}[b]{\linewidth}\raggedright
Random, with
\end{minipage} & \begin{minipage}[b]{\linewidth}\raggedright
Random, without
\end{minipage} & \begin{minipage}[b]{\linewidth}\raggedright
Grouped, with
\end{minipage} & \begin{minipage}[b]{\linewidth}\raggedright
Grouped, without
\end{minipage} \\
\midrule\noalign{}
Reduced standard & 28 & 0.931 / 0.037 & 0.315 / 0.133 & 0.870 / 0.047 & 0.229 / 0.136 \\
Standard & 2,753 & 0.816 / 0.069 & 0.808 / 0.070 & 0.720 / 0.088 & 0.715 / 0.089 \\
Reduced modified & 7 & 0.947 / 0.029 & 0.784 / 0.060 & 0.260 / 0.116 & 0.378 / 0.114 \\
Modified & 66 & 0.740 / 0.058 & $-$0.651 / 0.162 & 0.457 / 0.081 & $-$0.269 / 0.135 \\
\bottomrule\noalign{}
\end{tabular}
\end{table*}

}

\begin{figure*}[!tp]\centering
\includegraphics[width=\textwidth]{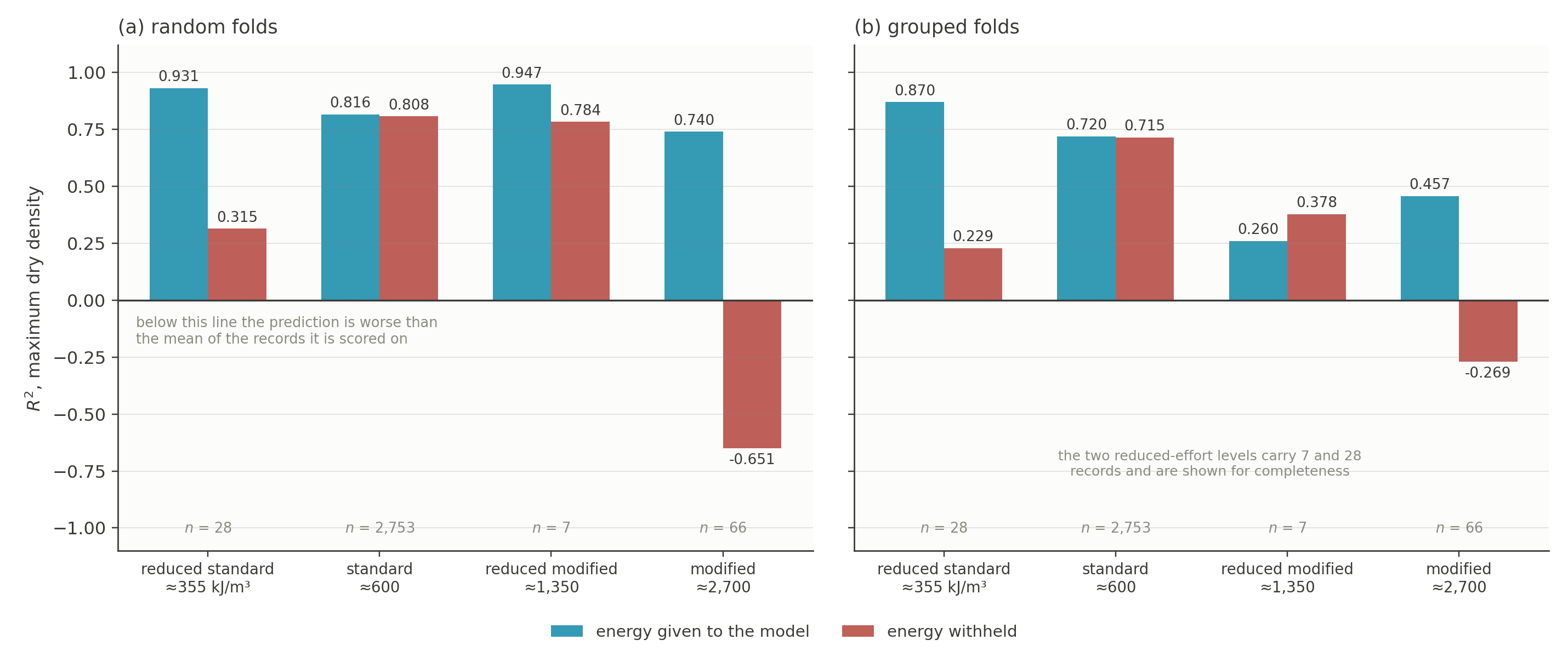}
\caption{R\textsuperscript{2} for density by compactive effort, with and without the energy input, gradient-boosted model, under (a) random and (b) grouped folds.}
\end{figure*}

The modified-Proctor row is the relevant one. Given the energy variable, the
model predicts those 66 records at R\textsuperscript{2} 0.740 under random folds. Deprived of it,
the prediction falls below the mean of the records it is scored on, to $-$0.651,
and the mean absolute error rises from 0.058 to 0.162 Mg/m\textsuperscript{3}. The model has no
means of knowing that those soils were compacted at 4.5 times the standard
effort, so it defaults to the standard-Proctor relationship that describes almost
all of its training data. The same happens less sharply under grouped folds. The
standard-Proctor row, 96.5 \% of the corpus, barely moves. That contrast is the
substance of the result. A variable can be worthless on average and decisive on
the stratum that requires it. The reduced-modified row rests on 7 records and
carries no weight.

Table 15 cannot be extended to the source-held-out design, for a reason that is a
property of the corpus rather than of the method. All 101 non-standard records
come from figshare 32955851, spread over 51 of its 83 groups. A grouped fold
therefore still contains non-standard records in training. A source-held-out fold
that removes that deposit removes every non-standard record with it, leaving the
energy variable present but untrained at three of its four levels. That is one
reason it is the hardest source of the six in Table 11.

The result does not depend on the precision with which those energies are
recorded, and cannot, because the variable takes four distinct values across
2,854 records. For the gradient-boosted model the exact conversion, the value
recomputed from the governing standard and the rounded SI convention all return
R\textsuperscript{2} 0.819 and 0.783, as does an ordered category carrying no magnitude. Four
indicator variables return 0.819 and 0.780. Compactive energy is therefore a
categorical variable carried as a number, and the numerical value functions as a
label.

\subsection{Closed-form expressions by symbolic regression}\label{closed-form-expressions-by-symbolic-regression}

Gradient boosting provides the accuracy of Table 10 but not an expression a
specification can quote. Symbolic regression was run to establish the cost of a
closed form. MDD and OMC are searched independently, as they are modelled
independently throughout. In the recommended form the two are then coupled
through the phase relation, so that admissibility is structural.

PySR {[}35,36{]} implements a multi-population evolutionary search over
expression trees, minimising mean-square loss against a complexity, and returning
the most accurate expression at every complexity rather than a single model. The
operator set is \(\{+, -, \times, \div, \sqrt{\cdot}, \log, (\cdot)^2,
(\cdot)^{-1}\}\). The reciprocal is included as a primitive because \(1/w_L\) is the
motif every density search rediscovers. Nested unary operators are forbidden. The
reported front comes from 250 iterations over 36 populations at maximum size 40
with parsimony \(2\times10^{-4}\). PySR cannot accept a missing value, so for this
search alone sand is dropped and the 134 non-plastic soils excluded. That leaves
2,720 records and six inputs with all four compactive energies intact. The search
runs on a random 80 \% of those records, 2,176 in total, and the front is scored
on the remaining 544.

The protection that partition provides is limited, as Section 4.3 states. The
0.692 and 0.658 quoted below are selection scores rather than scores on records
untouched by every step. The transfer claim of this paper does not rest on them.
It rests on the refitted figures below, where the structure is held fixed and
every constant is re-estimated inside each training fold.

\subsubsection{The recommended pair}\label{the-recommended-pair}

The search returns a front rather than a single expression, so one entry has to
be selected. Density is taken at complexity 20, on a front that is flat over a
wide range of length. Three symbols already reach R\textsuperscript{2} 0.613 and complexity 11
reaches 0.669. The entry at complexity 18 scores marginally higher, 0.693 against
0.692, but it carries no energy term, which Section 5.4 shows would be an unsound
omission. Complexity 20 retains \(\ln\bar{E}\) at a cost of 0.001. The optimum is
then recovered from the predicted density through the phase relation rather than
from a direct expression:

\begin{align}
\frac{\mathrm{MDD}}{\rho_w} &= 1.969 + 0.0100\,\frac{G_s^{\,2}}{w_P + I_P}
  - 0.3285\,F \nonumber\\
  &\quad - 0.8985\,w_P + 0.003285\,\ln\bar{E}
  \label{eq:mdd-symbolic}\\[6pt]
\mathrm{OMC} &= \frac{0.820}{G_s}
  \left(\frac{G_s\,\rho_w}{\mathrm{MDD}} - 1\right)
  \label{eq:omc-symbolic}
\end{align}

Every term is a pure number. The plastic limit, the plasticity index and the
fines content enter as the reported percentages divided by 100 \%. A soil at
\(w_P\) = 18 \% and \(I_P\) = 15 \% with 45 \% fines therefore gives \(w_P\) = 0.18,
\(I_P\) = 0.15 and \(F\) = 0.45, and OMC likewise returns a decimal fraction.
Specific gravity is already a ratio, density is divided by \(\rho_w\) = 1 Mg/m\textsuperscript{3},
and \(\bar{E} = E/E_0\) with \(E_0\) = 592.5 kJ/m\textsuperscript{3} the standard-Proctor reference, so
\(\ln\bar{E}\) is zero at standard effort.

Equation \eqref{eq:omc-symbolic} is the phase relation of Section 2.4 read in
reverse, and Figure 4 shows it. It is an exact identity, so the only estimated
quantity in it is the constant 0.820. At a fixed saturation the equation is a
curve in the MDD--OMC plane, one per specific gravity, lying under the
zero-air-voids bound by construction. The pair is therefore this paper's density
model composed with an identity rather than a second fit.

\begin{figure*}[!tp]\centering
\includegraphics[width=\textwidth]{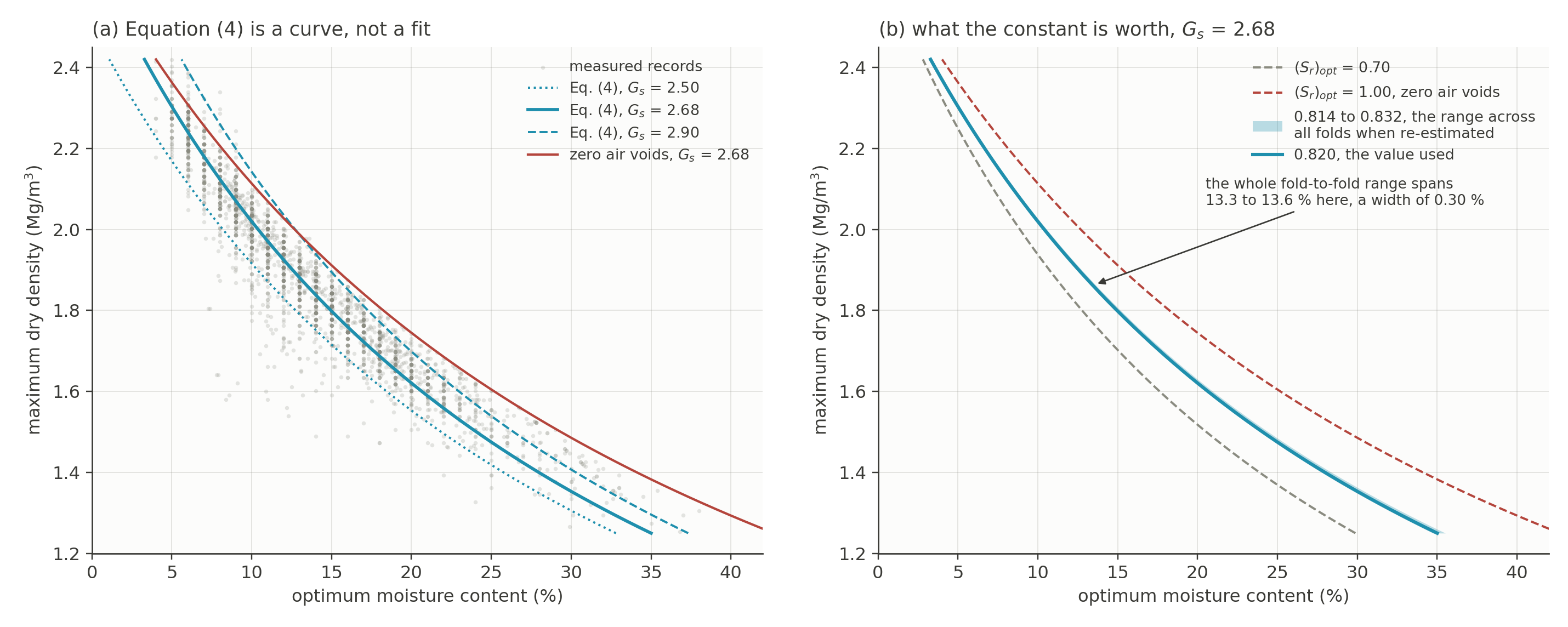}
\caption{Equation \eqref{eq:omc-symbolic}: (a) the identity at $(S_r)_{opt} = 0.820$ for three specific gravities, with the measured records behind it; (b) the fold-to-fold range of the constant against the span a different saturation would imply.}
\end{figure*}

The formulation is motivated by the optimum-saturation framework of Tatsuoka and
Gomes Correia {[}14{]} and is not taken from it. That framework relates density,
water content and saturation to one another. It does not predict MDD and OMC from
index properties, so Equation \eqref{eq:omc-symbolic} is not a Tatsuoka predictor
and is not scored as one anywhere in this paper. The constant is estimated here as
the mean measured saturation over the 2,720 records of the symbolic subset. That
it falls close to the average of about 0.82 reported for the standard-Proctor
compilation of Tatsuoka and Gomes Correia {[}14{]} is external corroboration rather
than the origin of the number. Re-estimated inside every fold it spans 0.814 to
0.832, worth 0.30 \% in water content at the median density, whereas a different
saturation would give a different curve entirely.

Two features of the density expression merit comment. Every sign is the physical
one and none was prompted. Density falls with fines, with the plastic limit and
with the plasticity index, and rises with specific gravity and with compactive
effort. The recurring grouping \(G_s^2/(w_P + I_P)\) is nonetheless empirical. It is
returned because density falls with both limits and one reciprocal costs fewer
symbols than two linear terms, and it should be read as a fitted grouping rather
than as a mechanism. The expressions also hold over the corpus on which they were
fitted and not beyond it. The 2,720 records span a plastic limit of 1.0 to 49.3 \%,
a plasticity index of 1.0 to 126.2 \%, a fines content of 1.5 to 100 \%, a specific
gravity of 2.30 to 2.94 and the four Proctor energies. Outside those ranges the
equations extrapolate, and the released predictor flags such an input.

Figure 5 shows the pair applied to the 544 held-out records. Panel (c) addresses
the claim in the title. The predicted pairs lie on a single constant-saturation
contour because Equation \eqref{eq:omc-symbolic} places them there, and none of
the 544 can cross the zero-air-voids bound for any input.

\begin{figure*}[!tp]\centering
\includegraphics[width=\textwidth]{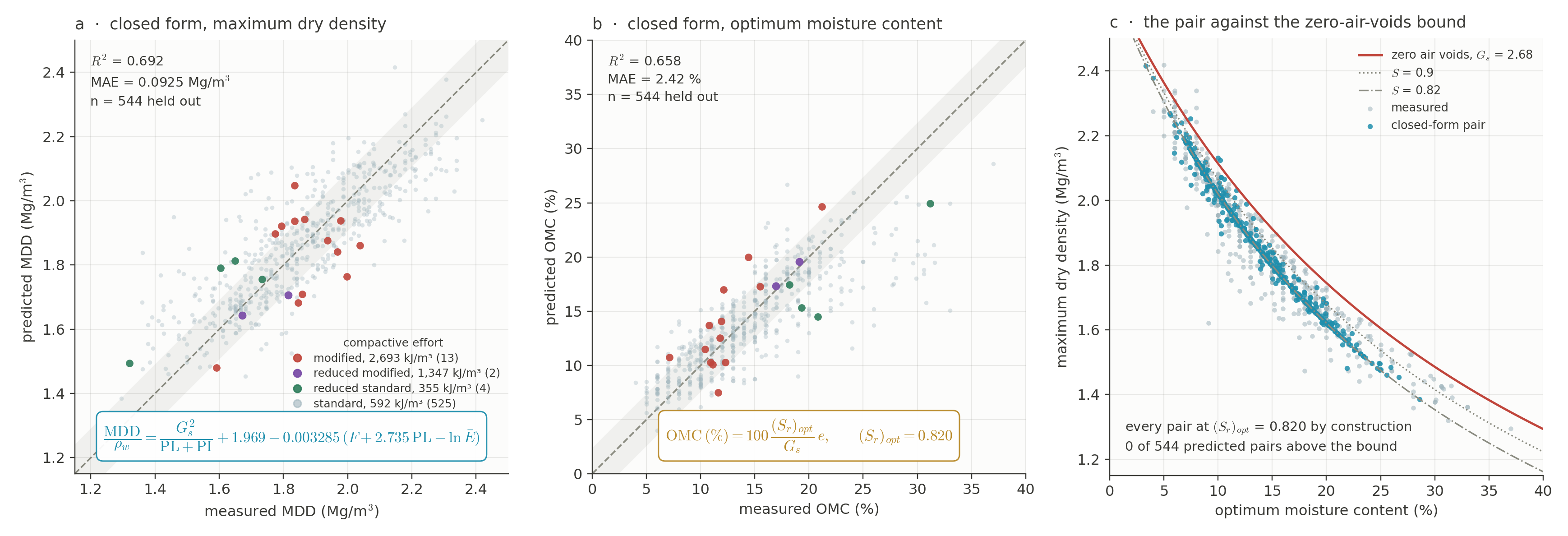}
\caption{The closed-form pair on the 544 held-out records: (a) density, Equation \eqref{eq:mdd-symbolic}; (b) the optimum recovered through Equation \eqref{eq:omc-symbolic}; (c) the predicted pairs against the zero-air-voids bound.}
\end{figure*}

\subsubsection{Why the coupling, and what it costs}\label{why-the-coupling-and-what-it-costs}

A second search was given the void ratio implied by
Equation \eqref{eq:mdd-symbolic} alongside every soil variable. It returned a
single term proportional to the void ratio and discarded the rest. That is worth
recording, because the search was free to introduce plasticity, gradation or
effort terms and did not. It returned \(\mathrm{OMC}\,(\%) = \lambda e\) with
\(\lambda \approx 30.70\), a constant \(\mathrm{OMC}/e\) and not a constant-saturation
form. Substituting it into the phase relation gives \((S_r)_{opt} = 0.307\,G_s\), so
the implied saturation rises with specific gravity from 0.706 at the corpus
minimum to 0.903 at its maximum.

Table 16 compares four ways of obtaining OMC from the same predicted density.
Every coefficient in it, including \((S_r)_{opt}\), is estimated within the training
fold. The density expression is common to all four rows, so the rows differ only
in the coupling.

{\def\LTcaptype{none} 
\begin{table*}[!tp]\centering
\caption*{Table 16 --- Four ways of obtaining OMC from the same predicted density, over the 2,720 records of the symbolic subset. \(n\) is the count of predicted pairs above the zero-air-voids line, out of 2,720. The common density expression returns R\textsuperscript{2} 0.710, 0.676 and 0.579 under the three designs.}
\begin{tabular}{@{}
>{\raggedright\arraybackslash}p{(\linewidth - 8\tabcolsep) * \real{0.6471}}
  >{\raggedleft\arraybackslash}p{(\linewidth - 8\tabcolsep) * \real{0.0882}}
  >{\raggedleft\arraybackslash}p{(\linewidth - 8\tabcolsep) * \real{0.0882}}
  >{\raggedleft\arraybackslash}p{(\linewidth - 8\tabcolsep) * \real{0.0882}}
  >{\raggedleft\arraybackslash}p{(\linewidth - 8\tabcolsep) * \real{0.0882}}@{}}
\toprule\noalign{}
\begin{minipage}[b]{\linewidth}\raggedright
Formulation for OMC
\end{minipage} & \begin{minipage}[b]{\linewidth}\raggedleft
OMC R\textsuperscript{2}
\end{minipage} & \begin{minipage}[b]{\linewidth}\raggedleft
MAE (\%)
\end{minipage} & \begin{minipage}[b]{\linewidth}\raggedleft
Pairs above ZAV, \(n\)
\end{minipage} & \begin{minipage}[b]{\linewidth}\raggedleft
max \(S\)
\end{minipage} \\
\midrule\noalign{}
\textbf{Random folds} & & & & \\
A --- independent expression & 0.678 & 2.34 & 75 & 1.29 \\
as first published, \(\mathrm{OMC} = \lambda e\) & 0.681 & 2.36 & 0 & 0.90 \\
\textbf{B --- phase relation, constant \((S_r)_{opt}\)} & \textbf{0.688} & \textbf{2.31} & \textbf{0} & \textbf{0.82} \\
C --- phase relation, predicted \((S_r)_{opt}\) & 0.700 & 2.31 & 0 & 0.96 \\
\textbf{Grouped folds} & & & & \\
A --- independent expression & 0.669 & 2.37 & 98 & 1.56 \\
as first published, \(\mathrm{OMC} = \lambda e\) & 0.657 & 2.45 & 0 & 0.90 \\
\textbf{B --- phase relation, constant \((S_r)_{opt}\)} & \textbf{0.666} & \textbf{2.40} & \textbf{0} & \textbf{0.82} \\
C --- phase relation, predicted \((S_r)_{opt}\) & 0.677 & 2.40 & 0 & 0.96 \\
\textbf{Source held out} & & & & \\
A --- independent expression & 0.659 & 2.45 & 81 & 132.7 \\
as first published, \(\mathrm{OMC} = \lambda e\) & 0.592 & 2.72 & 0 & 0.92 \\
\textbf{B --- phase relation, constant \((S_r)_{opt}\)} & \textbf{0.606} & \textbf{2.65} & \textbf{0} & \textbf{0.83} \\
C --- phase relation, predicted \((S_r)_{opt}\) & 0.619 & 2.65 & 0 & 0.96 \\
\bottomrule\noalign{}
\end{tabular}
\end{table*}

}

\emph{Bold marks the recommended formulation, not the highest R\textsuperscript{2}. Form C scores 0.012
higher in every design and is rejected on the grounds given below.}

Coupling secures admissibility, and it secures it completely. The independent
expression places 75 of 2,720 pairs above the zero-air-voids line under random
folds, 98 under grouped folds and 81 with a source held out. One of them reaches a
saturation of 132.7 in the last design, which is not a small error. All three
coupled forms produce none, under all three designs, at an R\textsuperscript{2} spread of 0.022
across the four rows.

Form B is recommended. It is bounded for any \(G_s\), whereas \(\lambda e\) is
admissible only because \(\lambda < G_s/100\) happens to hold for every soil
present, and it would fail above \(G_s \approx 3.26\). Letting the saturation itself
be predicted, form C, adds 0.012 at the cost of a second fitted model. Because
Equation \eqref{eq:omc-symbolic} requires only the predicted density and \(G_s\), it
composes with any density source, including the ensemble of Section 5.1.

Table 17 sets the closed form against the gradient-boosted model on the same
records and the same folds, so that the price of the closed form is a measurement
rather than an assertion.

{\def\LTcaptype{none} 
\begin{table*}[!tp]\centering
\caption*{Table 17 --- The closed-form pair against the ensemble, on the 2,720 records of the symbolic subset. Each cell gives R\textsuperscript{2} for density then water content.}
\begin{tabular}{@{}
>{\raggedright\arraybackslash}p{(\linewidth - 6\tabcolsep) * \real{0.5357}}
  >{\raggedright\arraybackslash}p{(\linewidth - 6\tabcolsep) * \real{0.1548}}
  >{\raggedright\arraybackslash}p{(\linewidth - 6\tabcolsep) * \real{0.1548}}
  >{\raggedright\arraybackslash}p{(\linewidth - 6\tabcolsep) * \real{0.1548}}@{}}
\toprule\noalign{}
\begin{minipage}[b]{\linewidth}\raggedright
Estimator
\end{minipage} & \begin{minipage}[b]{\linewidth}\raggedright
Random
\end{minipage} & \begin{minipage}[b]{\linewidth}\raggedright
Grouped
\end{minipage} & \begin{minipage}[b]{\linewidth}\raggedright
Source held out
\end{minipage} \\
\midrule\noalign{}
Equations \eqref{eq:mdd-symbolic} and \eqref{eq:omc-symbolic} & 0.710 / 0.688 & 0.676 / \textbf{0.666} & 0.579 / \textbf{0.606} \\
Gradient-boosted, five inputs (\(w_L\), \(w_P\), \(F\), \(\ln\bar{E}\), \(G_s\)) & 0.804 / 0.770 & 0.677 / 0.631 & 0.501 / 0.557 \\
Gradient-boosted, all seven inputs & \textbf{0.821} / \textbf{0.782} & \textbf{0.716} / 0.661 & \textbf{0.590} / 0.548 \\
\bottomrule\noalign{}
\end{tabular}
\end{table*}

}

\emph{Bold marks the highest value in each column and target.}

The closed form costs about 0.11 in R\textsuperscript{2} under random folds. That gap closes almost
entirely under grouped folds and reverses for water content with a source held
out, where the coupled expression beats both ensembles. An expression with five
coefficients has far less capacity to memorise a source than an ensemble, which
is what Table 17 shows.

Two limits apply to this section. The figures are confined to the 2,720 records
of the symbolic subset, so they are not those of Table 10. And the functional form
was selected on these data by a search that saw every source. What is established
is that the fitted constants transfer, not that the structure was found without
sight of the corpus. On these six inputs the closed forms are not recommended over
the ensemble for a soil outside the corpus. What holds without qualification is
that they cannot violate the zero-air-voids condition.

\subsection{The optimum degree of saturation}\label{the-optimum-degree-of-saturation}

A third symbolic search, targeting the degree of saturation at the compaction
peak, returned a constant. No expression up to size 26 exceeded R\textsuperscript{2} 0.05, because
little variation remains once the mean is removed. Figure 6(a) shows the
distribution. Over the 2,854 records \(S_{opt}\) has a mean of 0.815 and a
coefficient of variation of 11 \%. The distribution is tight and bounded above by
the zero-air-voids line. The quantity is that of Tatsuoka and Gomes Correia
{[}14{]}, and the contribution here is a measurement of it at scale rather than its
definition.

\begin{figure*}[!tp]\centering
\includegraphics[width=\textwidth]{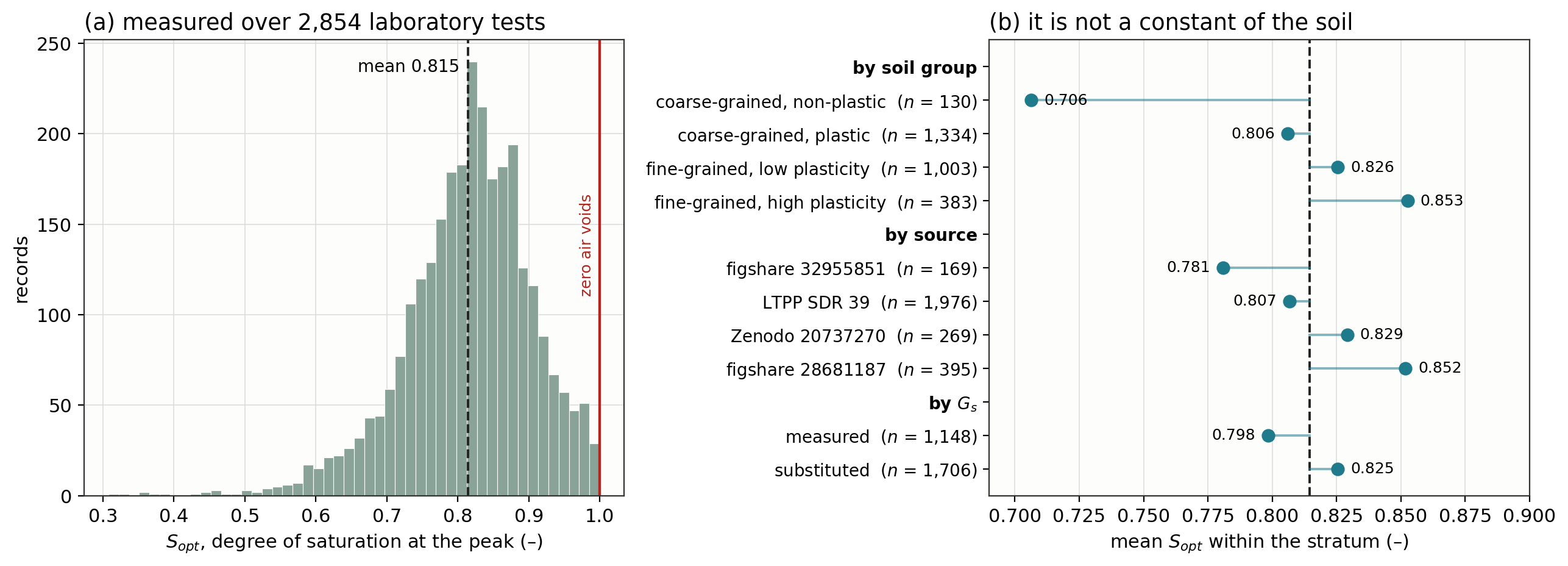}
\caption{The degree of saturation at the compaction peak: (a) its distribution over the 2,854 records, against the zero-air-voids bound; (b) its mean within each stratum, against the corpus mean.}
\end{figure*}

Table 18 gives the same quantity by stratum, and Figure 6(b) draws it. The
measurement does not support treating 0.815 as a constant of the soil.

{\def\LTcaptype{none} 
\begin{table*}[!tp]\centering
\caption*{Table 18 --- The optimum degree of saturation by stratum, 2,854 records. CV is the coefficient of variation. The corpus mean has a 95 \% interval of 0.811 to 0.818, or 0.802 to 0.824 when clustered on the 162 provenance groups. A fifth soil group, fine-grained non-plastic, holds 4 records and is omitted.}
\begin{tabular}{@{}
>{\raggedright\arraybackslash}p{(\linewidth - 8\tabcolsep) * \real{0.5294}}
  >{\raggedleft\arraybackslash}p{(\linewidth - 8\tabcolsep) * \real{0.0882}}
  >{\raggedleft\arraybackslash}p{(\linewidth - 8\tabcolsep) * \real{0.0882}}
  >{\raggedleft\arraybackslash}p{(\linewidth - 8\tabcolsep) * \real{0.0882}}
  >{\raggedright\arraybackslash}p{(\linewidth - 8\tabcolsep) * \real{0.2059}}@{}}
\toprule\noalign{}
\begin{minipage}[b]{\linewidth}\raggedright
Stratum
\end{minipage} & \begin{minipage}[b]{\linewidth}\raggedleft
\(n\)
\end{minipage} & \begin{minipage}[b]{\linewidth}\raggedleft
Mean
\end{minipage} & \begin{minipage}[b]{\linewidth}\raggedleft
CV (\%)
\end{minipage} & \begin{minipage}[b]{\linewidth}\raggedright
5th--95th percentile
\end{minipage} \\
\midrule\noalign{}
\textbf{Whole corpus} & 2,854 & \textbf{0.815} & 11.1 & 0.655--0.951 \\
\emph{By soil group} & & & & \\
coarse-grained, non-plastic & 130 & 0.706 & 13.4 & 0.576--0.887 \\
coarse-grained, plastic & 1,334 & 0.806 & 11.8 & 0.649--0.949 \\
fine-grained, low plasticity & 1,003 & 0.826 & 8.6 & 0.705--0.941 \\
fine-grained, high plasticity & 383 & 0.853 & 9.7 & 0.714--0.981 \\
\emph{By compactive effort} & & & & \\
reduced standard & 28 & 0.850 & 9.5 & 0.741--0.993 \\
standard & 2,753 & 0.815 & 10.9 & 0.659--0.951 \\
reduced modified & 7 & 0.892 & 5.5 & 0.835--0.956 \\
modified & 66 & 0.788 & 17.0 & 0.575--0.944 \\
\emph{By treatment of specific gravity} & & & & \\
measured \(G_s\) & 1,148 & 0.798 & 12.2 & 0.628--0.940 \\
substituted \(G_s\) & 1,706 & 0.825 & 10.1 & 0.682--0.958 \\
\emph{By source} & & & & \\
LTPP SDR 39 & 1,976 & 0.807 & 10.3 & 0.655--0.935 \\
figshare 28681187 & 395 & 0.852 & 7.5 & 0.747--0.958 \\
Zenodo 20737270 & 269 & 0.829 & 12.3 & 0.661--0.981 \\
figshare 32955851 & 169 & 0.781 & 19.6 & 0.459--0.974 \\
Zenodo 14251190 & 38 & 0.864 & 4.3 & 0.810--0.923 \\
Zenodo 19242689 & 7 & 0.906 & 7.7 & 0.817--0.980 \\
\bottomrule\noalign{}
\end{tabular}
\end{table*}

}

\emph{Bold marks the whole-corpus reference against which each stratum is read. No
value in this table is a performance figure.}

Soil group moves the mean systematically, from 0.706 for coarse-grained
non-plastic soils to 0.853 for fine-grained soils of high plasticity. That range
of 0.147 is more than six times the width of the cluster-robust interval on the
corpus mean. Source means and the treatment of specific gravity also move it, the
latter by 0.027 in the direction the substitution predicts. Compactive effort
moves it least, which is the one result that agrees with the earlier work without
qualification, although at 3.5 \% non-standard records these data do not test it
severely. Within-group variance accounts for 74 \% of the total and between-group
variance for 26 \%, so most of the scatter is not explained by provenance either.

What the measurement supports is a baseline rather than a constant. The corpus
figure of 0.815, with a fifth-to-ninety-fifth percentile spread of 0.655 to
0.951, is of the same order as the 64.1 \% to 92.3 \% reported across the nine soils
of Tatsuoka et al.~{[}15{]}, but is obtained from 2,854 laboratory records rather
than nine. The corpus mean is a defensible first estimate where no test exists. It
is not a substitute for determining \((S_r)_{opt}\) for the material at hand, which
is what a control procedure normalised on \(\Delta S_r\) requires.

\section{Limitations}\label{limitations}

Table 19 lists what the corpus cannot support. Each row states the limitation,
the evidence for it, and the claim it restricts. The paragraphs below expand the
three that most affect how the results should be used.

{\def\LTcaptype{none} 
\begin{table*}[!tp]\centering
\caption*{Table 19 --- What the corpus cannot support.}
\begin{tabular}{@{}
>{\raggedright\arraybackslash}p{(\linewidth - 4\tabcolsep) * \real{0.2979}}
  >{\raggedright\arraybackslash}p{(\linewidth - 4\tabcolsep) * \real{0.3511}}
  >{\raggedright\arraybackslash}p{(\linewidth - 4\tabcolsep) * \real{0.3511}}@{}}
\toprule\noalign{}
\begin{minipage}[b]{\linewidth}\raggedright
Limitation
\end{minipage} & \begin{minipage}[b]{\linewidth}\raggedright
Evidence
\end{minipage} & \begin{minipage}[b]{\linewidth}\raggedright
What it restricts
\end{minipage} \\
\midrule\noalign{}
Laboratory-only & no source reports a field density & every accuracy, closed form and \((S_r)_{opt}\) figure is a laboratory statement \\
Peak only & no public source retains curve points & the curve away from the optimum is not modelled \\
Thin energy coverage & 101 non-standard records, all from one deposit & the energy result of Table 15 cannot be tested source-out \\
Coarse source folds & six sources, one holding 69 \% & Table 11 confounds transfer with training-set size \\
Coarse grouping variable & an LTPP group is a state, not a laboratory & no laboratory-held-out evaluation is possible \\
\(w_L\) dropped & gains the learners 0.041, costs the closed form 0.108 & the closed forms would be better fitted on all three limits \\
Sand boundary not uniform & 1,976 records at 4.75 mm, 878 as published & the sand column mixes two definitions \\
\(G_s\) substituted on 59.8 \% & screen rate moves 19.8 \% to 8.2 \% over 2.60--2.75 & the 11.8 \% rejection figure, not the finding behind it \\
Structure selected on these data & the Pareto entry was chosen with held-out accuracy in view & the closed-form structure, not its constants \\
No independent test set & all results cross-validated within 2,854 records & every figure in this paper \\
\bottomrule\noalign{}
\end{tabular}
\end{table*}

}

\textbf{The corpus is laboratory-only.} The accuracies of Section 5, the closed forms
and the mean \((S_r)_{opt}\) of 0.815 all describe laboratory compaction. Whether
\((S_r)_{opt}\) near 0.82 also describes field-compacted lifts is not tested here
and cannot be tested on a corpus of this kind. The comparison with Tatsuoka and
Gomes Correia {[}14{]} is therefore one across different evidence bases.

\textbf{The grouping variable is as coarse as the sources allow.} An LTPP group is a
state rather than a laboratory. No source publishes the laboratory that performed
each test, so a strictly laboratory-held-out evaluation cannot be run on any
public corpus known to the authors. That is the single change to reporting
practice that would most improve work of this kind. The source-held-out design is
also coarse and cannot be otherwise. A single programme contributes 69 \% of the
records and the corpus has six sources, so each fold changes the training set
substantially. One per-source figure in Table 11 rests on 7 records.

\textbf{No independent test set exists.} All results are cross-validated within the
same 2,854 records, and the input set, the ablations and the constant of Section
5.5 were chosen after inspecting them. The source-held-out design is the closest
substitute available without new laboratory work, and it is not equivalent. A
held-out source is still one the authors selected, harmonised and screened under
the conventions of Section 2, whereas a genuinely new laboratory would also differ
in the mould, the oversize correction and the rammer. The degradation reported in
Table 10 is a lower bound on the cost of a new laboratory rather than an estimate
of it.

Two further rows deserve a number. Dropping the liquid limit was decided on the
learners and charged to the closed forms. It gains the gradient-boosted model
0.041 in density with a source held out (Table 13), but the symbolic search
performs worse without it. An earlier version whose search could use \(w_L\)
returned a density expression reaching R\textsuperscript{2} 0.687 with a source held out, against
0.579 for Equation \eqref{eq:mdd-symbolic}. A variable that adds nothing to a
model with capacity to spare can still be load-bearing for a simple expression, so
where the closed form is the priority it should be fitted on all three limits.
Specific gravity is nearly inert as an input but not as the denominator of the
screening rule. It also costs accuracy in the expected direction: against the
1,148 records reporting a measured value, the 1,706 substituted records lose 0.018
in density and 0.044 in water content, since \(G_s\) enters the optimum through the
phase relation and the density only weakly.

\section{Conclusions}\label{conclusions}

The primary contribution of this study is a corpus of laboratory compaction
tests. On it a model was built for maximum dry density and optimum moisture
content, and closed-form expressions were fitted for the same two parameters.

\emph{The corpus.}

\begin{enumerate}
\def\labelenumi{\arabic{enumi}.}
\item
  2,854 laboratory compaction tests were assembled from six public sources, five
  of them deposits meeting five stated retention conditions out of 618
  candidates (Tables 2 and 3). The corpus is an order of magnitude larger than
  the datasets behind the published correlations. It spans four Proctor energy
  levels and both sides of the coarse--fine boundary. Every record is the peak of
  a moisture--density curve determined to a Proctor method its source names, and
  the compactive energy is the standard the source states rather than a value
  inferred here. The corpus holds no field density, so every result reported here
  concerns laboratory compaction. It is released with per-record provenance under
  CC BY 4.0, together with the query strings, the screening decisions and the
  source adapters.
\item
  Provenance is what the corpus adds beyond size. Every record carries an
  identifier for the publication, site or deposit it came from, giving 162
  groups (Table 7). The two transfer-facing fold designs of Table 9 are built on
  that field, and neither can be run without it.
\item
  A material share of published compaction data is physically impossible, and
  the share depends on how specific gravity is treated. Screening removed 11.8 \%
  of harmonised records overall and 5.7 \% of those reporting a measured value.
  The rate runs from 19.8 \% to 8.2 \% as the substituted value moves from 2.60 to
  2.75 (Table 5). The zero-air-voids condition is recommended as a filter rather
  than as a diagnostic.
\item
  The optimum degree of saturation of Tatsuoka and Gomes Correia {[}14{]} is
  quantified across 2,854 laboratory tests at a mean of 0.815 and a coefficient
  of variation of 11 \%, with a fifth-to-ninety-fifth percentile range of 0.655 to
  0.951 (Table 18). It is a corpus-level baseline rather than a soil-independent
  constant, and a laboratory value rather than evidence of field invariance.
\end{enumerate}

\emph{The model.} Three learner classes were carried throughout and eleven
architectures evaluated, on identical folds and inputs.

\begin{enumerate}
\def\labelenumi{\arabic{enumi}.}
\setcounter{enumi}{4}
\item
  Under the random five-fold design used in this literature, a tabular
  foundation model predicts maximum dry density at R\textsuperscript{2} 0.824 and a mean absolute
  error of 0.066 Mg/m\textsuperscript{3}, and optimum moisture content at 0.784 and 1.87 \%.
  Gradient-boosted trees return 0.819 and 0.783 (Table 10).
\item
  Those figures measure interpolation within the corpus. Under that design
  96.6 \% of records share a provenance group with their training fold, and 40.1 \%
  share an exact input vector (Table 7). Grouped folds cost about 0.10 in R\textsuperscript{2} on
  both targets at an unchanged training-set size, and holding out a whole source
  costs more again. Every claim made here about a new laboratory rests on those
  two designs.
\item
  Under random folds the data rather than the capacity of the model limits
  accuracy. Eleven architectures agree closely (Table A1), and admitting records
  with largely missing gradation and energy costs 0.283 in density. Under
  source-held-out folds the learners cease to agree, and much of that spread is
  the missing-data encoding rather than the architecture. One gradient-boosted
  model moves from 0.392 to 0.597 in density when gaps are median-imputed with an
  indicator instead of passed to the learner. A comparison run on random folds
  therefore identifies neither the model nor the protocol that transfers.
\item
  The consistency limits and gradation are the two load-bearing input groups,
  each worth 0.096 to 0.213 in R\textsuperscript{2}. Sand alone, compactive energy and specific
  gravity never reach 0.030 (Table 12). Of the three consistency
  limits only two are independent, and \(w_P\) with \(I_P\) matches or exceeds all
  three in every design, by 0.041 in density with a source held out (Table 13).
\item
  Compactive energy is marginally negligible yet conditionally decisive.
  Omitting it drives modified-Proctor density prediction to R\textsuperscript{2} $-$0.651, far below
  the mean of those records (Table 15). Its effect is categorical, five encodings
  differing by at most 0.003. Only a multi-energy corpus can show this, which is
  why the four levels were kept at the cost of an incomplete sand column.
\end{enumerate}

\emph{The closed forms.}

\begin{enumerate}
\def\labelenumi{\arabic{enumi}.}
\setcounter{enumi}{9}
\tightlist
\item
  With the structure fixed and every constant re-estimated inside the training
  fold, the recommended pair reaches R\textsuperscript{2} 0.710 and 0.688 under random folds,
  0.676 and 0.666 under grouped folds, and 0.579 and 0.606 with a whole source
  held out (Table 17). Coupling separates the formulations on admissibility
  rather than on accuracy. An independent OMC expression places 75 to 98
  predicted pairs above the zero-air-voids line depending on the design, and
  every coupled form places none (Table 16). The closed forms are not
  recommended over the ensemble for a soil outside this corpus. What holds
  without qualification is that they cannot violate the zero-air-voids
  condition.
\end{enumerate}

\emph{What the corpus asks of the next dataset.}

\begin{enumerate}
\def\labelenumi{\arabic{enumi}.}
\setcounter{enumi}{10}
\tightlist
\item
  The practical recommendation concerns compaction databases rather than
  models. The compaction standard should be recorded. Specific gravity should be
  reported as measured rather than assumed. An identifier should be carried for
  the laboratory and study each test came from. Without the last, no evaluation
  can test transfer to a new laboratory. The corpus released here carries all
  three fields, and its acquisition protocol and screening rule are released
  with it.
\end{enumerate}

\section*{CRediT authorship contribution statement}

\textbf{Sompote Youwai:} Conceptualization, Methodology, Software, Formal analysis,
Data curation, Writing -- original draft, Writing -- review \& editing,
Visualization, Supervision, Funding acquisition. \textbf{Chana Phutthananon:}
Methodology, Validation, Writing -- review \& editing. \textbf{Warat Kongkitkul:}
Conceptualization, Methodology, Writing -- review \& editing, Supervision.

\section*{Declaration of competing interest}

The authors declare that they have no known competing financial interests or
personal relationships that could have appeared to influence the work reported
in this paper.

\section*{Declaration of generative AI and AI-assisted technologies in the writing process}

During the preparation of this work the authors used a large language model
(Claude, Anthropic) to assist with editing prose for concision, with
restructuring results into tabular form, and with routine analysis scripting.
After using this tool the authors reviewed and edited the content as needed and
take full responsibility for the content of the published article. No text,
figure, table or numerical result was accepted without verification against the
released data and analysis code. Generative AI was not used to generate,
interpret or alter any measurement, and is not credited as an author.

\section*{Acknowledgements}

The authors thank the depositors of the six datasets listed in Table 3, whose
decision to publish their records under open licences made this corpus possible.

\section*{Data availability}

The corpus, the fitted models, the command-line predictor and all analysis code
are released at \url{https://github.com/Sompote/open_compaction} under CC BY 4.0, at
version v2.2.0, with a checksum manifest, a data dictionary and a validation
script. The repository is public and requires no account to read. The dataset is
one flat file, one row per laboratory compaction test, carrying the six inputs,
the two targets, the provenance identifier and the compaction standard. The
liquid limit and a flag separating measured from substituted specific gravity
are held in a wider file beside it. The code reproduces every quantity reported
here, including the record-level out-of-fold predictions under all three fold
designs, so any stratification can be recomputed without refitting. Analyses
used Python 3.11 with xgboost 2.1.4, scikit-learn 1.5.1, PySR 0.19.4 and version
6.4.1 of the PyPI package \texttt{tabpfn}. All seeds were fixed. The symbolic search is
multi-threaded and is therefore not bit-reproducible.
{[}DOI TO BE INSERTED ON DEPOSIT.{]}

\section*{References}

{[}1{]} Proctor RR. Fundamental principles of soil compaction. Eng News-Rec 1933;111:245--8.

{[}2{]} ASTM D698-12(2021). Standard test methods for laboratory compaction characteristics of soil using standard effort (12,400 ft-lbf/ft\textsuperscript{3} (600 kN$\cdot$m/m\textsuperscript{3})). West Conshohocken, PA: ASTM International; 2021.

{[}3{]} ASTM D1557-12(2021). Standard test methods for laboratory compaction characteristics of soil using modified effort (56,000 ft-lbf/ft\textsuperscript{3} (2,700 kN$\cdot$m/m\textsuperscript{3})). West Conshohocken, PA: ASTM International; 2021.

{[}4{]} AASHTO T99. Standard method of test for moisture--density relations of soils using a 2.5-kg (5.5-lb) rammer and a 305-mm (12-in.) drop. Washington, DC: American Association of State Highway and Transportation Officials.

{[}5{]} AASHTO T180. Standard method of test for moisture--density relations of soils using a 4.54-kg (10-lb) rammer and a 457-mm (18-in.) drop. Washington, DC: American Association of State Highway and Transportation Officials.

{[}6{]} Lambe TW. The structure of compacted clay. J Soil Mech Found Div 1958;84(SM2). \url{https://doi.org/10.1061/JSFEAQ.0000114}.

{[}7{]} Seed HB, Chan CK. Structure and strength characteristics of compacted clays. J Soil Mech Found Div 1959;85(SM5):87--128. \url{https://doi.org/10.1061/JSFEAQ.0000229}.

{[}8{]} Mitchell JK, Soga K. Fundamentals of soil behavior. 3rd ed.~Hoboken, NJ: John Wiley \& Sons; 2005.

{[}9{]} Daniel DE, Benson CH. Water content--density criteria for compacted soil liners. J Geotech Eng 1990;116:1811--30. \url{https://doi.org/10.1061/(ASCE)0733-9410(1990)116:12(1811)}.

{[}10{]} Baek S-H, Kim J, Park K, Choi H. Field study on intelligent compaction for compaction quality control of subgrade bases. Can Geotech J 2025;62:1--14. \url{https://doi.org/10.1139/cgj-2024-0300}.

{[}11{]} Li H, Sego DC. Equation for complete compaction curve of fine-grained soils and its applications. In: Constructing and controlling compaction of earth fills, ASTM STP 1384. West Conshohocken, PA: ASTM International; 2000, p.~113--25. \url{https://doi.org/10.1520/STP15277S}.

{[}12{]} Horpibulsuk S, Katkan W, Apichatvullop A. An approach for assessment of compaction curves of fine grained soils at various energies using a one point test. Soils Found 2008;48:115--25. \url{https://doi.org/10.3208/sandf.48.115}.

{[}13{]} Horpibulsuk S, Katkan W, Naramitkornburee A. Modified Ohio's curves: a rapid estimation of compaction curves for coarse- and fine-grained soils. Geotech Test J 2009;32:64--75. \url{https://doi.org/10.1520/GTJ101659}.

{[}14{]} Tatsuoka F, Gomes Correia A. Importance of controlling the degree of saturation in soil compaction linked to soil structure design. Transp Geotech 2018;17:3--23. \url{https://doi.org/10.1016/j.trgeo.2018.06.004}.

{[}15{]} Tatsuoka F, Yoshida T, Nagai H, Tomita Y, Latimer R, Kikuchi Y, Koseki J. Soil compaction control by monitoring compacted states based on soil stiffness indices. Soils Found 2026;66:101709. \url{https://doi.org/10.1016/j.sandf.2025.101709}.

{[}16{]} Sridharan A, Nagaraj HB. Plastic limit and compaction characteristics of fine-grained soils. Proc Inst Civ Eng Ground Improv 2005;9:17--22. \url{https://doi.org/10.1680/grim.2005.9.1.17}.

{[}17{]} Gurtug Y, Sridharan A. Compaction behaviour and prediction of its characteristics of fine grained soils with particular reference to compaction energy. Soils Found 2004;44:27--36. \url{https://doi.org/10.3208/sandf.44.5_27}.

{[}18{]} Nagaraj HB, Reesha B, Sravan MV, Suresh MR. Correlation of compaction characteristics of natural soils with modified plastic limit. Transp Geotech 2015;2:65--77. \url{https://doi.org/10.1016/j.trgeo.2014.09.002}.

{[}19{]} Sharma B, Sridharan A. Static method to determine compaction characteristics of fine-grained soils. Geotech Test J 2016;39:1048--55. \url{https://doi.org/10.1520/GTJ20150221}.

{[}20{]} Vinod P, Sridharan A. Toughness limit: a useful index property for prediction of compaction parameters of fine-grained soils at any rational compactive effort. Indian Geotech J 2016;47:107--14. \url{https://doi.org/10.1007/s40098-016-0194-6}.

{[}21{]} Pandian NS, Nagaraj TS, Manoj M. Re-examination of compaction characteristics of fine-grained soils. Géotechnique 1997;47:363--6. \url{https://doi.org/10.1680/geot.1997.47.2.363}.

{[}22{]} Blotz LR, Benson CH, Boutwell GP. Estimating optimum water content and maximum dry unit weight for compacted clays. J Geotech Geoenviron Eng 1998;124:907--12. \url{https://doi.org/10.1061/(ASCE)1090-0241(1998)124:9(907)}.

{[}23{]} Günaydın O. Estimation of soil compaction parameters by using statistical analyses and artificial neural networks. Environ Geol 2009;57:203--15. \url{https://doi.org/10.1007/s00254-008-1300-6}.

{[}24{]} Khatti J, Grover KS. Prediction of compaction parameters for fine-grained soil: critical comparison of the deep learning and standalone models. J Rock Mech Geotech Eng 2023;15:3010--38. \url{https://doi.org/10.1016/j.jrmge.2022.12.034}.

{[}25{]} Rehman ZU, Khalid U, Ijaz N, Ijaz Z. Big data-driven global modeling of cohesive soil compaction across conceptual and arbitrary energies through machine learning. Transp Geotech 2025;50:101470. \url{https://doi.org/10.1016/j.trgeo.2024.101470}.

{[}26{]} Verma G, Sivapullaiah PV. Prediction of compaction parameters for fine-grained and coarse-grained soils: a review. Int J Geotech Eng 2019;14:970--7. \url{https://doi.org/10.1080/19386362.2019.1595301}.

{[}27{]} Page MJ, McKenzie JE, Bossuyt PM, Boutron I, Hoffmann TC, Mulrow CD, et al.~The PRISMA 2020 statement: an updated guideline for reporting systematic reviews. BMJ 2021;372:n71. \url{https://doi.org/10.1136/bmj.n71}.

{[}28{]} Rethlefsen ML, Kirtley S, Waffenschmidt S, Ayala AP, Moher D, Page MJ, Koffel JB. PRISMA-S: an extension to the PRISMA statement for reporting literature searches in systematic reviews. Syst Rev 2021;10:39. \url{https://doi.org/10.1186/s13643-020-01542-z}.

{[}29{]} Federal Highway Administration. Introduction to the LTPP information management system (IMS). Report FHWA-RD-03-088. McLean, VA: Federal Highway Administration; 2003. \url{https://www.fhwa.dot.gov/publications/research/infrastructure/pavements/ltpp/reports/03088/} (accessed August 15, 2026).

{[}30{]} Federal Highway Administration. Long-term pavement performance standard data release 39, material test volume. InfoPave; 2025. \url{https://infopave.fhwa.dot.gov/Data/StandardDataRelease} (accessed August 15, 2026).

{[}31{]} Soranzo C. Dataset for ``Machine learning predictions on an extensive geotechnical database''. Zenodo; 2024. \url{https://doi.org/10.5281/zenodo.14251190}.

{[}32{]} ASTM D2487-17(2025). Standard practice for classification of soils for engineering purposes (Unified Soil Classification System). West Conshohocken, PA: ASTM International; 2025.

{[}33{]} ASTM D4318-17e1. Standard test methods for liquid limit, plastic limit, and plasticity index of soils. West Conshohocken, PA: ASTM International; 2017.

{[}34{]} ASTM D854-14. Standard test methods for specific gravity of soil solids by water pycnometer. West Conshohocken, PA: ASTM International; 2014. Withdrawn in 2023 and replaced by ASTM D854-23, Standard test methods for specific gravity of soil solids by the water displacement method.

{[}35{]} Cranmer M. PySR: high-performance symbolic regression in Python and Julia, v0.19.4 {[}software{]}. Zenodo; 2023. \url{https://doi.org/10.5281/zenodo.4041458}.

{[}36{]} Cranmer M. Interpretable machine learning for science with PySR and SymbolicRegression.jl. arXiv:2305.01582; 2023. \url{https://doi.org/10.48550/arXiv.2305.01582}.

{[}37{]} Hollmann N, Müller S, Purucker L, Krishnakumar A, Körfer M, Hoo SB, Schirrmeister RT, Hutter F. Accurate predictions on small data with a tabular foundation model. Nature 2025;637:319--26. \url{https://doi.org/10.1038/s41586-024-08328-6}.

{[}38{]} Müller S, Hollmann N, Arango SP, Grabocka J, Hutter F. Transformers can do Bayesian inference. In: International Conference on Learning Representations (ICLR); 2022. \url{https://doi.org/10.48550/arXiv.2112.10510}.

{[}39{]} Roberts DR, Bahn V, Ciuti S, Boyce MS, Elith J, Guillera-Arroita G, et al.~Cross-validation strategies for data with temporal, spatial, hierarchical, or phylogenetic structure. Ecography 2017;40:913--29. \url{https://doi.org/10.1111/ecog.02881}.

\appendices

\section{Architectures from the compaction literature}\label{architectures-from-the-compaction-literature}

The learners of Table 10 are not those most used in this literature. Neural
networks, support vector machines, random forests and stacked hybrids recur
across it {[}23--26{]}. None of those studies releases its fitted parameters, so
none can be applied to these records as published. The architectures can be
refitted on identical folds and inputs, which Table A1 reports. None accepts a
missing value, so all share one preprocessing: training-fold median, a
was-missing indicator per gapped column, and standardisation on training-fold
statistics. None is tuned on these records, so the table compares model families
at fixed effort rather than at fixed tuning budget.

{\def\LTcaptype{none} 
\begin{table*}[!tp]\centering
\caption*{Table A1 --- Architectures from the compaction literature, refitted on identical folds and inputs. Each cell gives R\textsuperscript{2} for density then water content. The second column names a study that applied that family to this task, not the source of any hyperparameter. Two rows are variants added here. The last three rows are the learners of Table 10.}
\begin{tabular}{@{}
>{\raggedright\arraybackslash}p{(\linewidth - 8\tabcolsep) * \real{0.3820}}
  >{\raggedright\arraybackslash}p{(\linewidth - 8\tabcolsep) * \real{0.1798}}
  >{\raggedright\arraybackslash}p{(\linewidth - 8\tabcolsep) * \real{0.1461}}
  >{\raggedright\arraybackslash}p{(\linewidth - 8\tabcolsep) * \real{0.1461}}
  >{\raggedright\arraybackslash}p{(\linewidth - 8\tabcolsep) * \real{0.1461}}@{}}
\toprule\noalign{}
\begin{minipage}[b]{\linewidth}\raggedright
Architecture
\end{minipage} & \begin{minipage}[b]{\linewidth}\raggedright
Used by
\end{minipage} & \begin{minipage}[b]{\linewidth}\raggedright
Random
\end{minipage} & \begin{minipage}[b]{\linewidth}\raggedright
Grouped
\end{minipage} & \begin{minipage}[b]{\linewidth}\raggedright
Source held out
\end{minipage} \\
\midrule\noalign{}
multiple linear regression & {[}23{]} & 0.679 / 0.696 & 0.630 / 0.675 & 0.513 / 0.588 \\
shallow ANN, one hidden layer & {[}23{]} & 0.768 / 0.738 & 0.715 / 0.701 & 0.506 / 0.550 \\
multilayer perceptron, three layers & {[}24{]} & 0.791 / 0.756 & 0.717 / 0.690 & 0.577 / 0.555 \\
support vector regression, RBF & {[}24{]} & 0.788 / 0.743 & 0.709 / 0.650 & 0.471 / 0.496 \\
random forest & {[}25{]} & 0.814 / 0.772 & 0.719 / 0.677 & 0.579 / 0.588 \\
extremely randomised trees & added here & 0.820 / 0.782 & 0.715 / 0.681 & 0.581 / 0.569 \\
gradient boosting, scikit-learn & {[}25{]} & 0.803 / 0.774 & 0.724 / 0.695 & \textbf{0.606} / 0.588 \\
stacked hybrid, RF + SVR + ANN & added here & 0.813 / 0.779 & \textbf{0.730} / 0.687 & 0.587 / 0.604 \\
\emph{XGBoost, this pipeline} & {[}25{]} & 0.817 / 0.782 & 0.719 / 0.681 & 0.597 / 0.597 \\
\emph{XGBoost, gaps passed directly} & --- & 0.819 / 0.783 & 0.722 / 0.681 & 0.392 / 0.225 \\
\emph{TabPFN, gaps passed directly} & {[}37{]} & \textbf{0.824} / \textbf{0.784} & 0.727 / \textbf{0.696} & 0.520 / \textbf{0.614} \\
\bottomrule\noalign{}
\end{tabular}
\end{table*}

}

\emph{Bold marks the highest value in each design column, separately for density and
water content.}

Under random folds the table confirms the reading of Section 5.1. The seven tree
and ensemble architectures span 0.021 in density, and eight of the eleven learners
fall within 0.036 of the best. The exceptions are the linear model, the shallow
network and the support vector machine. The conclusion that the data rather than
the learner limits what is reached under this design therefore rests on eleven
architectures rather than three. Under grouped folds the pattern is the same at a
lower level.

The source-held-out column is the informative one. Scikit-learn's gradient
boosting reaches 0.606 for density and the stacked hybrid 0.587, against 0.392 for
the gradient-boosted model of Table 10 and 0.520 for the foundation model. The
cause is not the architecture, and the anchor row establishes this. The same
\texttt{XGBRegressor}, with the same hyperparameters and the same folds, reaches 0.597
and 0.597 when the gaps are median-imputed with a was-missing indicator, and 0.392
and 0.225 when they are passed to the learner directly. That is a difference of
0.205 in R\textsuperscript{2} for density from the encoding alone. A gradient-boosted tree learns a
default branch for a missing value from the sources it trains on, and that branch
does not transfer to a source it has not seen, whereas an explicit indicator does.
Under random folds the two encodings are indistinguishable, at 0.819 and 0.817, so
the design this literature reports would not reveal the difference. A random split
therefore identifies neither the model that transfers nor the missing-data
protocol that transfers. The linear model no longer collapses either, returning
0.513 and 0.588 on the six inputs where on all three limits it returned $-$51 and
$-$101, so removing one near-collinear column accounts for the whole of that
difference.

\end{document}